\documentclass[letterpaper,12pt,leqno]{article}
\usepackage{paper}
\usepackage[font=small, labelfont=bf]{caption}
\usepackage{placeins}
\usepackage{amsmath}
\usepackage{epigraph}
\usepackage{mathrsfs}
\usepackage{rotating}
\usepackage{tabularx}
\usepackage{listings}
\usepackage{lipsum}
\usepackage{subcaption}
\usepackage{cleveref}
\usepackage{mathtools}
\usepackage[export]{adjustbox}
\usepackage{biblatex}
\usepackage{titling}
\hypersetup{pdftitle={On the Efficieny of Convolutional Neural Networks}}

\newcommand{\R}{\mathbb{R}}
\DeclareMathOperator{\ffn}{FFN}

\DeclarePairedDelimiter\floor{\lfloor}{\rfloor}

\begin{document}
\title{On the Efficiency of \protect\\ Convolutional Neural Networks}
\author{Andrew Lavin
\thanks{Andrew Lavin: Phantom AI. andrew@phantom.ai}}
\date{May 21, 2024}
\begin{titlepage}
  \maketitle
  \begin{figure}[h]
    \includegraphics[scale=0.59]{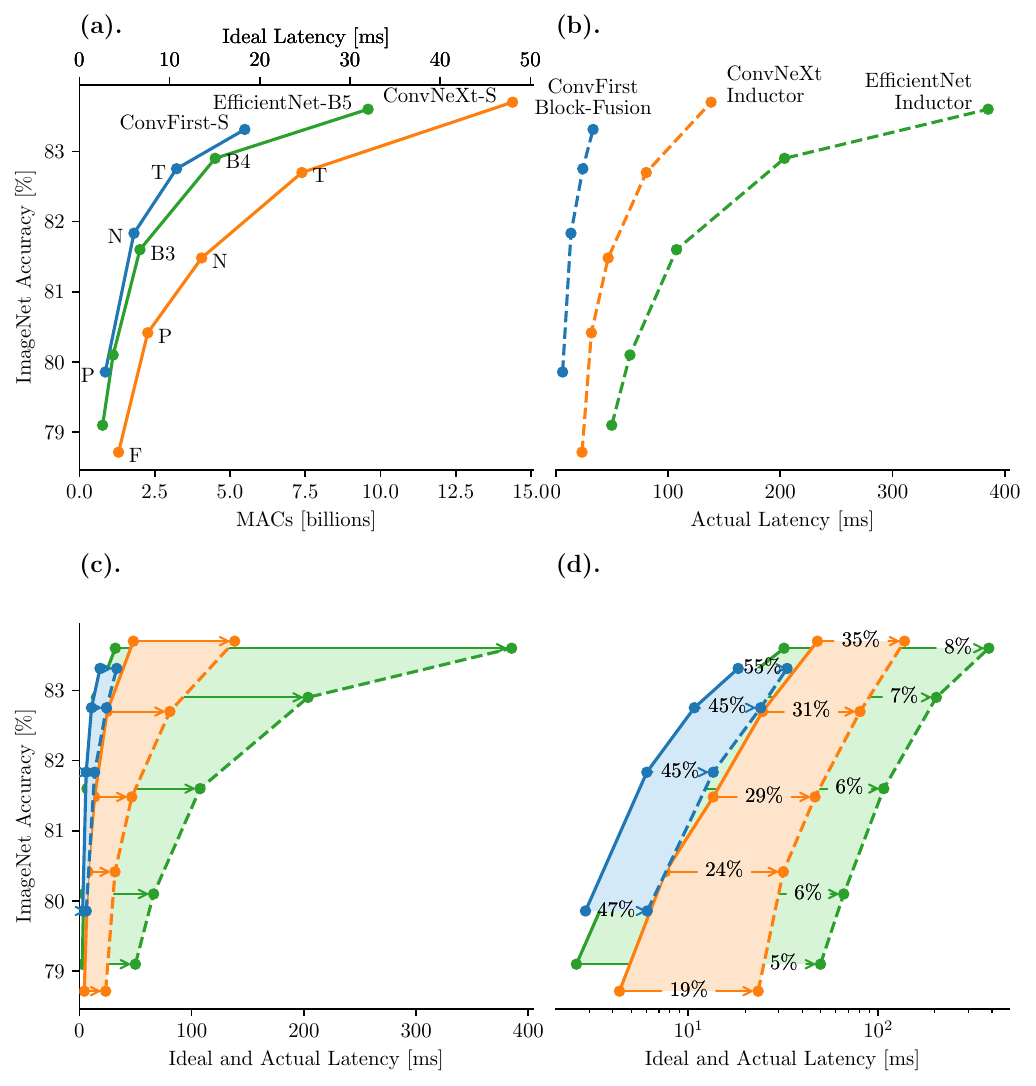}
    \caption*{\textbf{Efficiency gap plots} compare ideal and actual latency and quantify the difference as computational efficiency.
      \textbf{(a).} Dividing arithmetic complexity (MACs) by peak arithmetic throughput yields the \textit{ideal latency}.
      \textbf{(b).} Inference time on the GPU is the \textit{actual latency}.
      \textbf{(c).} The ideal latency of ConvNeXt is longer than the actual latency of ConvFirst.
      \textbf{(d).} Low computational efficiency causes a wide gap between ideal and actual latency. EfficientNet ranges from 5\% -- 8\% and has a wide gap. Our ConvFirst model with block-fusion kernels ranges from 47\% -- 55\% and has a narrow gap.
      We used an NVIDIA A5000 GPU with \lstinline[language=C]!float16! and batch size $128$.
    }
  \end{figure}
  \FloatBarrier
  \pagebreak
  \begin{abstract}
Since the breakthrough performance of AlexNet in 2012, convolutional neural networks (convnets) have grown into
extremely powerful vision models. Deep learning researchers have used convnets to perform vision tasks with accuracy that was unachievable a decade ago. Yet computer scientists make computational efficiency their primary objective.
They seek algorithms that use the fewest operations and the least communication.
Confronted with the immense computation that convnets use, deep learning researchers also became interested in efficiency.
They replaced expensive \lstinline[language=C]!conv2d! layers
with sequences of degenerate layers that have less arithmetic complexity.
They used network architecture search to find the model hyperparameters that
yield the greatest accuracy with the fewest operations. However, the engineers who deployed these efficient convnets soon realized that they were slower than the previous generation,
despite using fewer operations. Many reverted to older models that ran faster.
Hence researchers switched the objective of their search from arithmetic complexity to latency and produced
a new wave of models that performed better. Paradoxically, these models also used more operations. Skepticism grew among researchers and engineers alike about the relevance of arithmetic complexity.

Contrary to the prevailing view that latency and arithmetic complexity are irreconcilable, a simple
formula relates both through computational efficiency. This insight enabled us to co-optimize the separate
factors that determine latency. We observed that the degenerate \lstinline[language=C]!conv2d! layers that produce the best accuracy--complexity trade-off also use significant memory resources and have low computational efficiency.
We devised \textit{block fusion} algorithms to implement all the layers of a residual block in a single kernel, thereby creating temporal locality, avoiding communication, and
reducing workspace size. Our ConvFirst model with block-fusion kernels has less arithmetic complexity and greater computational efficiency than baseline models and kernels. ConvFirst ran approximately
four times as fast as ConvNeXt at equal accuracy on the ImageNet-1K classification task.

On the journey to this result, we created novel tools. We made the \textit{efficiency gap} plot to show accuracy, arithmetic complexity, latency, and computational efficiency
on the same axes. The plot illustrates how computational efficiency distorts model efficiency to produce latency.
We created \textit{waterline analysis} to extend roofline analysis to a sequence of parallel kernels.
We devised the \textit{tensor machine}, a simple abstract computer for tensor programs, and used it to plan
block-fusion kernels. Our unified approach to convnet efficiency envisions a new era of models and kernels that achieve greater accuracy at lower cost.

\end{abstract}
\end{titlepage}
\tableofcontents
\section{Introduction}\label{s:introduction}

The \lstinline[language=C]!conv2d! layer is the basic building block of convolutional neural networks (convnets). It accounts for most of a convnet's computation, so naturally it has been the focus of efforts to make convnets more efficient.
On the one hand, computer scientists improved the computational efficiency of the \lstinline[language=C]!conv2d! layer by examining its arithmetic complexity and data movement and applying this understanding to create more efficient algorithms and hardware. On the other hand, deep learning researchers reduced the arithmetic complexity
of the traditional \lstinline[language=C]!conv2d! layer by factoring it into a sequence of degenerate layers that perform fewer total operations.

Computer scientists improved the speed of first-wave convnets, while deep learning researchers created a second wave that used fewer operations to achieve the same accuracy. In other words, the original convnets benefited from improved \textit{computational efficiency} while the new ones had greater \textit{model efficiency}.
In practice, the second-wave models suffered from significantly worse computational efficiency that undermined their superior model efficiency.

Deep learning researchers responded in different ways to the efficiency paradox posed by the second wave.
Some employed network architecture search and found the convnet models with
the lowest latency on one or more software and hardware platforms. Others
exploited correlations between model hyperparameters and computational
efficiency and discovered that convnets could be made faster simply by making them wider (i.e., increasing the number of channels in every layer). Some optimized hardware architecture
and memory placement algorithms to make new processors that ran existing software more efficiently.

The result of these efforts was a third wave of convnets and accelerators that
were suboptimal in various ways. Despite their improved computational efficiency,
third-wave convnets often had worse model efficiency than second-wave
convnets. Third-wave accelerators performed with greater computational efficiency on all
models, while devoting much larger die area to on-chip memory (SRAM). These
accelerators had less peak arithmetic throughput per square millimeter than ones
that used less SRAM, effectively increasing the cost of an arithmetic operation.

We take a new approach that elevates the importance of computation. We find the simple formula
that relates model efficiency, computational efficiency, and latency, and plot all of these
quantities on the same axes, thereby revealing the efficiency gap that explains the convnet efficiency problem.
We extend roofline analysis to calculate the attainable latency of a sequence of parallel kernels; and we develop
simple plots that reveal the latency and operational intensity of individual layers. We use a simple abstract
tensor machine to express new algorithms and use it to design memory-efficient block-fusion kernels that greatly increase the attainable efficiency.
We contribute a new convnet that is co-optimized for model efficiency and computational efficiency, train it, and benchmark it with CUDA kernels that implement our block-fusion algorithms.
The results show greater model efficiency than our second-wave baseline and greater computational efficiency than the third-wave baseline. The combination of
efficient model and kernels yields significantly lower latency at equal levels of accuracy.

We hope our approach contributes to a new wave of convnets, software, and hardware that achieves significantly greater
efficiency.

\section{Related Work}

AlexNet~\cite{krizhevsky2012imagenet} began the modern era of convnets with its breakthrough performance on the ImageNet~\cite{deng2009imagenet} image classification benchmark.
Since then, researchers have published a series of innovative models, introducing new layers
that improved model efficiency. Network-in-Network used point-wise ($1 \times 1$) convolutions~\cite{lin2013network}. ResNet contributed the residual block and bottleneck block, enabling
deeper models and reducing the number of operations per activation~\cite{he2016deep}. Squeeze \&
Excitation networks introduced a lightweight, channel-wise attention mechanism
that increased accuracy while adding few additional
operations~\cite{hu2018squeeze}. MobileNetV2 introduced the inverted residual
block with depth-wise convolutions and a memory-efficient algorithm for
computing it~\cite{sandler2018mobilenetv2}. EfficientNet combined all of these ideas into
a scalable model that shattered the state-of-the-art for model efficiency over a wide range of scales~\cite{tan2019efficientnet}.

Early convnets used \lstinline[language=C]!conv2d! layers with $3 \times 3$ kernels or larger~\cite[\dots]{lecun1998gradient,krizhevsky2012imagenet,sermanet2013overfeat,simonyan2014very},
and early works on computational efficiency
focused on these layers. A straightforward implementation achieved $96\%$ computational efficiency on
an NVIDIA GPU, because of the layer's large operational intensity~\cite{lavin2015maxdnn}.
Subsequent works focused on fast algorithms
that reduced the arithmetic complexity of the convnet
block~\cite{mathieu2013fast, vasilache2014fast, cong2014minimizing}, culminating
in GPU kernels exceeding 100\% effective computational
efficiency~\cite{lavin2016fast}.

Recent convnets used point-wise~\cite{lin2013network} and 
depth-wise~\cite{chollet2017xception, howard2017mobilenets} convolutions. These degenerate \lstinline[language=C]!conv2d! layers have less
arithmetic complexity 
and increased model efficiency. Simultaneously, deep learning accelerators began
to use matrix multiply accelerators that dramatically increased the peak
arithmetic throughput~\cite{jouppi2017datacenter, nvidiav100}. Decreasing operational intensity of the
layers and increasing op:byte ratio of the accelerators put tremendous pressure
on the software that runs the convnets. The result was that EfficientNet~\cite{tan2019efficientnet} had
poor computational efficiency on a contemporary GPU using the best available software.

Several papers improved the performance of modern convnets by
using inference latency as the model cost instead of arithmetic complexity~\cite[\dots]{tan2019mnasnet, howard2019searching, dollar2021fast, brock2021high, tan2021efficientnetv2}.
These papers measured latency using popular deep learning frameworks and inference engines without modification.

Yang et al. optimized the data-flow, blocking, and memory hierarchy of convnet
accelerator hardware~\cite{yang2020interstellar}. Zhang et al. co-optimized the size of matrix multiply accelerators, on-chip memory capacity, and the memory placement of weights and activations
tensors~\cite{zhang2022full}. Both these works assumed layer-by-layer execution, and
both increased the on-chip memory capacity to avoid storing activations
to DRAM between layers.

Radosavovic et al.~\cite{radosavovic2020designing} observed that the number of
activations correlates with inference latency and searched for models with fewer
activations. Dollar et al.~\cite{dollar2021fast} used a scaling rule that
creates wider models, motivated by the observations that wider models have more operations per
activation and hence greater computational efficiency.

Other works computed a small number of tiles in one layer before computing the
next layer's tiles that depend on them~\cite{alwani2016fused,
  goetschalckx2019breaking}. This strategy created a depth-first execution of
the convnet that reduced the size of the workspace needed to hold activations,
making it possible to keep activations in a small on-chip memory. Depth-first
execution has long been used to exploit spatial locality in image
processing~\cite{ragan2013halide} and stencil programs~\cite{de2021stencilflow}. It cannot
traverse a layer that performs a global spatial reduction, such as Squeeze \& Excitation~\cite{hu2018squeeze}; nor
does it exploit the channels dimension of a convnet. Depth-first execution complements
our block-fusion strategy (see Section \ref{s:block-fusion}).

PyTorch 2 added the TorchInductor compiler backend that performs trivial layer
fusions and accelerates inference~\cite{ansel2024pytorch}. We used TorchInductor
as our baseline inference engine.

Our recent work~\cite{lavin2021pie,lavin2023mbconv} developed GPU kernels that compute all the layers of the
FusedMBConv~\cite{efficientnet_edgetpu} or MBConv~\cite{sandler2018mobilenetv2, tan2019efficientnet} block in a single
kernel. These block-fusion kernels exploit temporal locality to reduce workspace size and avoid DRAM
memory transfers. This paper extends those results.

Following the success of the transformer architecture for sequence
modeling~\cite{vaswani2017attention}, several works produced vision transformers~\cite[\dots]{dosovitskiy2020image, dai2021coatnet, liu2021swin, tu2022maxvit, dehghani2023scaling}. Vision transformers were scaled to very
large models with high accuracy. Smith et al.~\cite{smith2023convnets} found that convnets scale as
well as vision transformers with a similar training
budget. The focus of this paper is convnets, but the methodology we developed applies equally well to transformers and convnet-transformer hybrids.

\section{Efficiency Gap: How Computational Efficiency Distorts Model Efficiency}

\setlength{\epigraphwidth}{0.8\textwidth}

\epigraph{In physical science a first essential step in the direction of learning any subject is to find principles of numerical reckoning and methods for practicably measuring some quality connected with it. I often say that when you can measure what you are speaking about and express it in numbers you know something about it; but when you cannot measure it, when you cannot express it in numbers, your knowledge is of a meagre and unsatisfactory kind: it may be the beginning of knowledge, but you have scarcely, in your thoughts, advanced to the stage of \textit{science}, whatever the matter may be.}{\textit{William Thomson, 1st Baron Kelvin, 1883}~\cite{kelvin1889popular}}

In order to understand efficiency, we must first learn how to measure it. In this section we define various measures of efficiency. Some will already be known to the reader, others are borrowed from the field of high-performance computing, and one is novel. Where often-used metrics were not precisely named in previous literature, our nomenclature makes deliberate and unambiguous choices. Together our metrics describe the relationship between model and algorithm and illustrate how each contributes to performance.

After we have measured the different kinds of efficiency, we will develop a new kind of plot to visualize their relationships.

\subsection{Measuring Efficiency}

\textit{Model efficiency} $\mathscr{E}_m(n)$ measures accuracy as a
function of the number of mathematical operations $n$ the model performs. This metric
expresses the trade-off between the quality of the result (the accuracy) and the
quantity of work (computation). Thus it measures the efficiency with which the
model achieves the desired result. The accuracy of a model should increase as it scales to a larger number of operations, so we
assume that the model efficiency function is monotonically increasing.

Model efficiency is not bounded by the physics of computation. It says nothing
about the rate at which computation can be performed, so it cannot predict
latency.

There is no theory of machine learning that predicts the accuracy of a trained model. Thus model efficiency is, for now,
a purely empirical quantity. One measures model efficiency by instantiating a family of
models at different scales ($n$), training each model instance on the desired
task, and measuring the achieved accuracy.

\textit{Peak arithmetic throughput} $\mathscr{R}_i$ measures the fastest rate at which a processor can perform computation.
It represents the \textit{ideal} arithmetic throughput. Hardware manufacturers
report peak arithmetic throughput in units of operations per second (OP/s or
OPS). The unit TOPS indicates one trillion ($10^{12}$) operations per second. One multiply
and add (i.e., one multiply-accumulate or MAC) is considered to be $2$ OPs.
Almost all of the computation performed by neural networks is multiply-add
operations, so this definition of peak arithmetic throughput is useful for
understanding ideal performance.

\textit{Ideal latency}
\begin{equation}
  t_i(n) = \frac{n}{\mathscr{R}_i}
  \label{e:ideal-latency}
\end{equation}
is the fastest possible response time any model with $n$ operations could theoretically achieve on a processor with peak arithmetic throughput $\mathscr{R}_i$.

\textit{Ideal efficiency}
\begin{equation}
  \mathscr{E}_i(\frac{n}{\mathscr{R}_i}) = \mathscr{E}_m(n)
  \label{e:ei}
\end{equation}
is the greatest possible accuracy  a model can achieve for a given latency.

\textit{Actual arithmetic throughput}
\begin{equation}
  \mathscr{R}_a(n) = \frac{n}{t_a(n)}
\end{equation}
is the rate at which the processor and software perform the model's
operations, where $t_a(n)$ is the \textit{actual latency} measured for the model
with $n$ operations. Achieving high arithmetic throughput is a physics problem,
bounded by the physical behavior of the processor's arithmetic units, memory
systems, and the wires that connect them, and optimized by the algorithm that
schedules the utilization of these resources. In principle one could better
organize the processor and algorithm to achieve greater arithmetic throughput
for a given model.

\textit{Actual efficiency}
\begin{equation}
  \mathscr{E}_a\Big(\frac{n}{\mathscr{R}_a(n)}\Big) = \mathscr{E}_m(n)
  \label{e:ea}
\end{equation}
is the accuracy that the model, processor, and software achieve for a given latency.

Combining Equations \eqref{e:ei} and \eqref{e:ea},
\begin{equation}
  \mathscr{E}_m(n) = \mathscr{E}_i\Big(\frac{n}{\mathscr{R}_i}\Big) = \mathscr{E}_a\Big(\frac{n}{\mathscr{R}_a(n)}\Big)
  \label{e:3eff}
\end{equation}
we observe that ideal and actual efficiency are both horizontal dilations of the model efficiency function.

Alternatively, we can write
\begin{equation}
  \mathscr{E}_i(t) = \mathscr{E}_a\Big(\frac{t}{\mathscr{C}}\Big)
\end{equation}
where
\begin{equation}
\mathscr{C}(n) = \frac{\mathscr{R}_a(n)}{\mathscr{R}_i}
\label{e:comp-eff}
\end{equation}
is \textit{computational efficiency}, the ratio of actual to peak arithmetic throughput. Thus computational efficiency expands the ideal efficiency function to produce actual efficiency.

We can also write the relationship in terms of latencies as
\begin{equation}
t_a(n) = \frac{t_i(n)}{\mathscr{C}(n)}
\end{equation}
Using logarithmic scale we get
\begin{equation}
  \log\big(t_a(n)\big) = \log\big(t_i(n)\big) - \log\big(\mathscr{C}(n)\big)
  \label{e:log-gap}
\end{equation}
so that $- \log\big(\mathscr{C}(n)\big)$ is the amount by which the logarithm of the latency is increased because of imperfect computational efficiency. In an accuracy versus latency plot using logarithmic scale on the x-axis, the actual efficiency function is produced by shifting the ideal efficiency function by $- \log\big(\mathscr{C}(n)\big)$ at every model size $n$. We refer to the width of this latency shift as the \textit{efficiency gap}.

\subsection{Visualizing Efficiency}

\begin{figure}[h]
  \includegraphics[width=\textwidth]{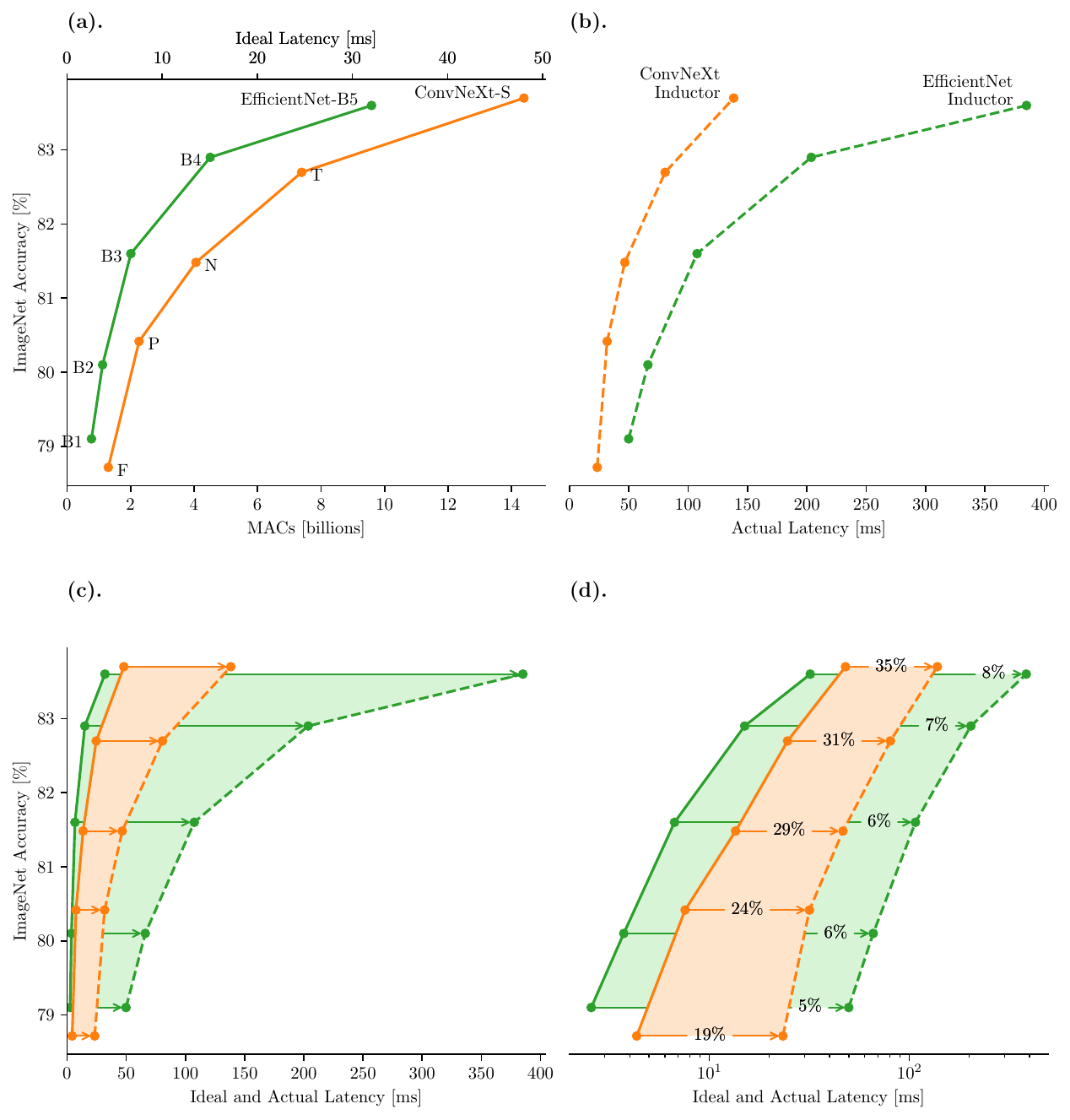}
  \caption{\textbf{Efficiency gap plots} illustrate the difference between ideal and actual performance for different models and software. They help us understand the separate contributions of model efficiency and computational efficiency. This figure shows ImageNet-1K classification accuracy and latency for EfficientNet and ConvNeXt using an NVIDIA Ampere A5000 GPU with $76.7$ TFLOP/s peak arithmetic throughput running PyTorch Inductor software. Batch size equals $128$.
    \textbf{(a).} \textit{Model efficiency} measures accuracy as a function of the number of multiply-accumulate operations (MACs) performed. Dividing MACs by peak arithmetic throughput of the processor yields \textit{ideal latency}, the lowest possible latency for the number of operations.
    \textbf{(b).} \textit{Actual efficiency} measures accuracy versus \textit{actual latency} for a combination of model and software.
    \textbf{(c).} Overlaying model and actual efficiency graphs reveals the \textit{efficiency gap}, the offset between ideal and actual latency.
    \textbf{(d).} Measured with a logarithmic scale on the latency axis, the (negative) width of the efficiency gap equals the logarithm of \textit{computational efficiency}, the ratio between actual and ideal performance. EfficientNet's poor computational efficiency creates a wide efficiency gap, resulting in longer latency than ConvNeXt, despite superior model efficiency.
  }
  \label{f:ce-gap}
\end{figure}

The previous section unified the concepts of model efficiency, latency,
and computational efficiency. Now we visualize them with the
\textit{efficiency gap plot} in Figure~\ref{f:ce-gap}. We introduce it by way of comparison between two
well known convnet models.

EfficientNet~\cite{tan2019efficientnet} set a new state-of-the-art in convnet model efficiency, using fewer operations than existing models while achieving higher accuracy. In practice, however, it ran slower than many older models on GPUs with available inference engine software.

ConvNeXt~\cite{liu2022convnet} promised to modernize and simplify convnets; and to provide a new baseline for comparison to vision transformer (ViT) models~\cite{dosovitskiy2020image}. ConvNeXt was soon recognized for its speed, and it served as a practical alternative to EfficientNet.

Figure~\ref{f:ce-gap}.a. shows model efficiency $\mathscr{E}_m$ for both
networks, with model size (MACs) on the x-axis and ImageNet classification
accuracy (\%) on the y-axis. This type of plot is well known from seminal papers
addressing the issue of model efficiency, including MobileNet
~\cite{howard2017mobilenets}, MobileNetV2~\cite{sandler2018mobilenetv2}, and
EfficientNet~\cite{tan2019efficientnet}. We observed that EfficientNet has better model efficiency than
ConvNeXt because it achieves greater accuracy with fewer operations (MACs).

We added a second x-axis to Figure \ref{f:ce-gap}.a. to show ideal latency
(milliseconds) and computed it with Equation~\eqref{e:ideal-latency}. Ideal
latency builds a bridge between model efficiency and actual latency in
subsequent plots.

We used peak arithmetic throughput $\mathscr{R}_i = 76.7$ for our NVIDIA Ampere
A5000 GPU (GA102~\cite{nvidia2022ga102}) with base clock frequency 1.17 GHz,
which is less than the boost clock frequency of 1.695 GHz. We also set the
memory clock to 1.25 GHz, which is less than the peak 2.0 GHz. The clock
reductions make the GPU performance stable during benchmarking while keeping the
op:byte ratio close to that which would be achieved at full clock speed. We used
a batch of 128 images, which is a typical workload for benchmarks on large GPUs.

Recognizing that arithmetic complexity (MACs) is an imperfect predictor of
latency, researchers switched to latency as the primary optimization
objective~\cite{tan2019mnasnet}. Figure \ref{f:ce-gap}.b. plots the accuracy of
EfficientNet and ConvNeXt versus their latency measured with PyTorch Inductor
software on the A5000 GPU. ConvNeXt has better actual efficiency than
EfficientNet in this experiment because it achieves greater accuracy at lower
latency.

The dissonance between model efficiency and latency has vexed deep
learning practitioners in recent years. One might be excited by a paper that
claims extraordinary model efficiency only to be frustrated by the high latency
of the deployed model. Skepticism has grown about the relevance of model efficiency.

Researchers found a reasonable approach: include both model efficiency and
latency plots when reporting results~\cite{dehghani2021efficiency}. We go a step further by plotting both metrics on the
same axes and showing the relationship between them.

Figure \ref{f:ce-gap}.c. combines Figures \ref{f:ce-gap}.a. and \ref{f:ce-gap}.b. by joining the ideal latency axis with the actual latency axis. Both axes have units of milliseconds, so they can simply be merged. The model efficiency and actual efficiency curves appear side-by-side on the same axes. For each model sample, we plot the accuracy on the y-axis and the ideal and actual latencies on the x-axis. A line between the ideal and actual latencies shows the width of the efficiency gap.

Figure \ref{f:ce-gap}.d. repeats the same plot using logarithmic scale on the x-axis. We calculated the computational efficiency using Equation \eqref{e:comp-eff} for each model sample and labeled the efficiency gaps with the corresponding computational efficiency.

Figure \ref{f:ce-gap}.d. shows that large differences between model efficiency and actual latency are caused by low computational efficiency. EfficientNet's computational efficiency ranges from 5\% to 8\% in our experiments, and its efficiency gap is much wider than its model efficiency advantage over ConvNeXt. Therefore, computational efficiency can be the dominant factor contributing to latency.

Also of interest is the fact that ConvNeXt's computational efficiency, at $35\%$ or less, is unimpressive in absolute terms, despite being relatively greater than EfficientNet's.

These plots help us understand the factors that contribute to the performance
difference between EfficientNet and ConvNeXt. They show that poor
computational efficiency undermines great model efficiency. They document in precise
analytical detail the source of the deep learning practitioner's frustration
with model efficiency.

These plots also suggest that the middle-road on which ConvNeXt travels is suboptimal,
as neither its model efficiency nor its computational efficiency is great. Best
performance requires co-optimization of model efficiency and computational efficiency.
Is it possible to excel at both? To answer this question, we must first
understand computational efficiency.

\FloatBarrier

\section{Waterline: A Simple Performance Model for a Sequence of Parallel Kernels}
\label{s:waterline}
Williams et al.~\cite{williams2009roofline} introduced the \textit{roofline} model to capture the idea that modern multi-processors have an increasingly high ratio of peak arithmetic throughput to DRAM memory bandwidth, and this imbalance presents the first hurdle that a parallel kernel must clear in order to be computationally efficient. 

We find that the roofline model explains most of the performance problems with contemporary inference engine software and use it as the foundation for a simple performance model that predicts the efficiency of a sequence of parallel kernels.

We adopt the formulation of the roofline model found in NVIDIA's \textit{GPU Performance Background User’s Guide}~\cite{gpuperf}.
\textit{Op:byte} measures the ratio of peak arithmetic throughput to DRAM memory bandwidth for a particular computer.
\textit{Operational intensity} measures the ratio of operations performed (OPs) to bytes transferred between DRAM and the processor for a parallel kernel~\cite{williams2009roofline}.
We count a single multiply-accumulate operation as two OPs~\cite{matmulback}.

If the operational intensity of a kernel is less than the op:byte ratio of the computer, then the kernel's performance will be limited by the DRAM memory bandwidth. Therefore, the roofline model says that the maximum attainable arithmetic throughput for a parallel kernel is
\begin{equation}
\mathscr{R}_{\max} = \min(\mathscr{R}, \mathscr{B} \frac{n}{b})
\end{equation}
where ${\mathscr{R}}$ is the peak arithmetic throughput, $\mathscr{B}$ is the peak DRAM memory bandwidth, $n$ is the number of operations, $b$ is the number of bytes transferred, and $\frac{n}{b}$ is the operational intensity~\cite{williams2009roofline}.

This formula implies that the minimum attainable latency for a parallel kernel is
\begin{equation}
t_{\min} = \frac{n}{\min(\mathscr{R}, \mathscr{B} \frac{n}{b})}
\end{equation}

The minimum attainable latency for a sequence of parallel kernels $i$ is
\begin{equation}
  T_{\min} = \sum_i{t_i} = \sum_i{\frac{n_i}{\min(\mathscr{R}, \mathscr{B} \frac{n_i}{b_i})}}
  \label{e:total-attainable-latency}
\end{equation}

The minimum attainable latency implies the maximum computational efficiency
\begin{equation}
  \mathscr{C}_{\max} = \frac{N}{T_{\min} \mathscr{R}}
  \label{e:max-comp-eff}
\end{equation}
where
\begin{equation}
N = \sum_i n_i
\end{equation}
is the total number operations across all kernels.

We visualize Equation \eqref{e:total-attainable-latency} as a sequence of kernels, side-by-side on a time axis, each of them reaching upwards with height equal to its operational intensity. If a kernel reaches
above the processor's op:byte \textit{waterline}, then it is compute bound, and can attain maximum performance. Otherwise, the ``underwater'' kernel is memory bound, and its performance is limited
by low operational intensity. Figure~\ref{f:waterline-baseline-models} plots the waterline performance for a number of baseline models.

In Appendix \ref{a:amdahls-roofline} we derive an extension to Amdah's Law using waterline analysis.

\subsection{Waterline Analysis of Conv2d Layers}

Convolutional neural networks are composed of \lstinline[language=C]!conv2d! layers. In this section we examine the function computed by \lstinline[language=C]!conv2d! and analyze its operational intensity.

\subsubsection{Conv2d and its Degenerate Cases}

The \lstinline[language=C]!conv2d! layer is defined by
\begin{equation}
  Y_{nhwk} = \sum_r^R \sum_s^S \sum_c^C A_{krsc} X_{n(h+r)(w+s)c} + b
  \label{f:conv2d}
\end{equation}
for input tensor $X \in \R^{N \times H \times W \times C}$, weights tensor $A\in \R^{K \times R \times S \times C}$, bias vector $b \in \R^{K}$, and output
tensor $Y \in \R^{N \times H \times W \times K}$; where $N$ is the batch size
(i.e., the number of images to be processed in parallel), $C$ is the number of
input channels, $H \times W$ is the size of the input and output feature maps, $K$ is the number
of output channels, and $R \times S$ is the size of the convolution kernel. Typically,
$X$ is padded (implicitly with $\floor{\frac{R}{2}}$ and $\floor{\frac{S}{2}}$ zeros on the top-bottom and left-right borders, respectively) to support
the receptive field of the convolution kernel.

One understands Equation \eqref{f:conv2d} as a convolution nested inside of a
matrix multiplication, with matrices $A \in \R^{K \times C}$ and $X \in \R^{C \times N}$,
matrix elements $w_{kc} \in \R^{R \times S}$ and $x_{cn} \in \R^{H \times W}$,
and element multiplication implemented by 2d
convolution~\cite{cong2014minimizing}. Conversely, it is a
matrix multiplication nested inside of a convolution, where the operands are 2d
images, the image elements are matrices, and element multiplication is implemented by
matrix multiplication~\cite{cong2014minimizing}. These formulations lend
themselves to arithmetic complexity reduction using fast matrix multiplication
and fast convolution algorithms~\cite{cong2014minimizing, lavin2016fast}.
The fast algorithms are most efficient when all the tensor dimensions are
large, because the cost of the algorithms' transforms are amortized over the
size of the tensor dimensions~\cite{cong2014minimizing, lavin2016fast}. Fast
algorithms can still be efficient for small tensor dimensions with specialized
hardware acceleration, especially if the transforms only use
additions and arithmetic shift operations.

\begin{figure}[h]
  \begin{subfigure}[b]{.3\textwidth}
    \includegraphics[scale=0.5,center]{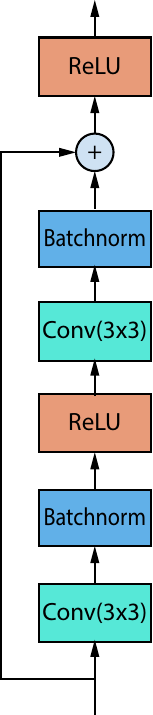}
    \caption{ResNet Block}
    \label{f:resnet-block}
  \end{subfigure}
  \begin{subfigure}[b]{.3\textwidth}
    \includegraphics[scale=0.5,center]{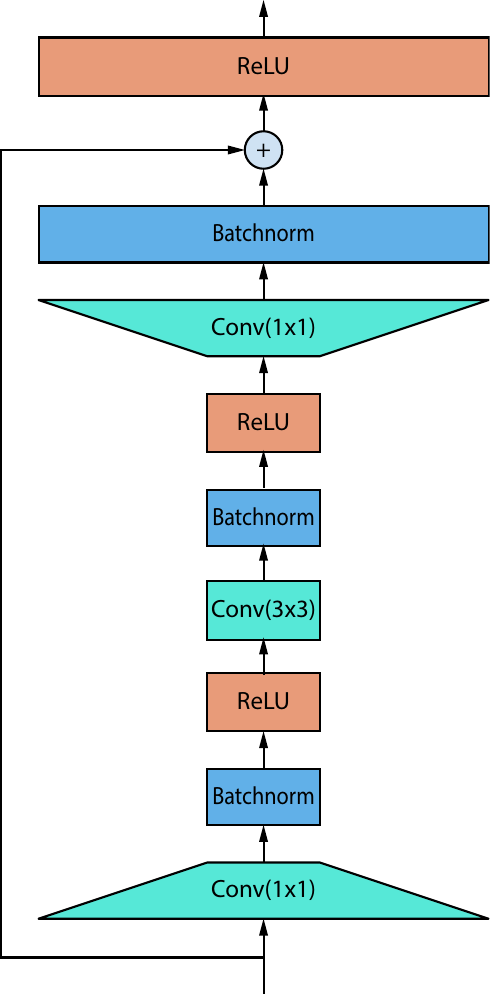}
    \caption{ResNet Bottleneck Block}
    \label{f:resnet-bottleneck-block}
  \end{subfigure}
  \begin{subfigure}[b]{.3\textwidth}
    \includegraphics[scale=0.5,center]{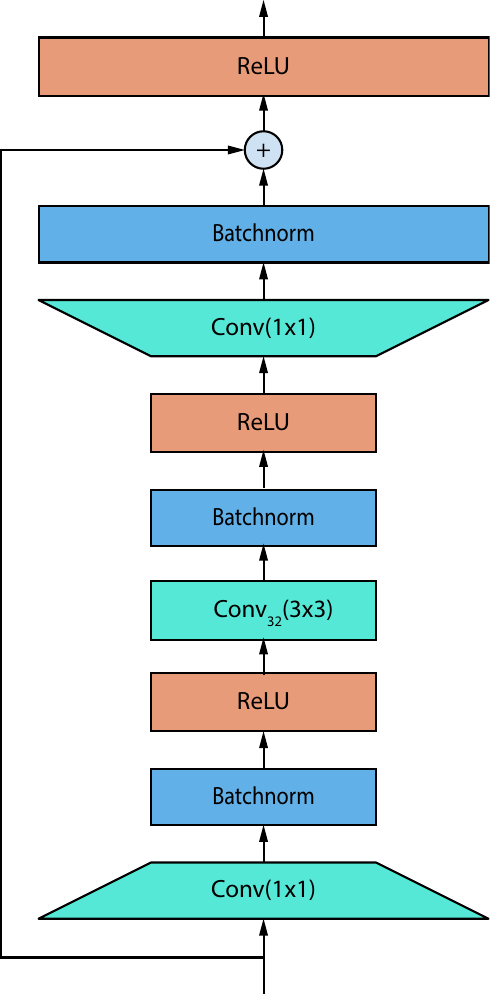}
    \caption{ResNeXt Bottleneck Block}
    \label{f:resnext-bottleneck-block}
  \end{subfigure}
  \vspace{4mm}
  \\
  \begin{subfigure}[b]{.45\textwidth}
    \includegraphics[scale=0.40,center]{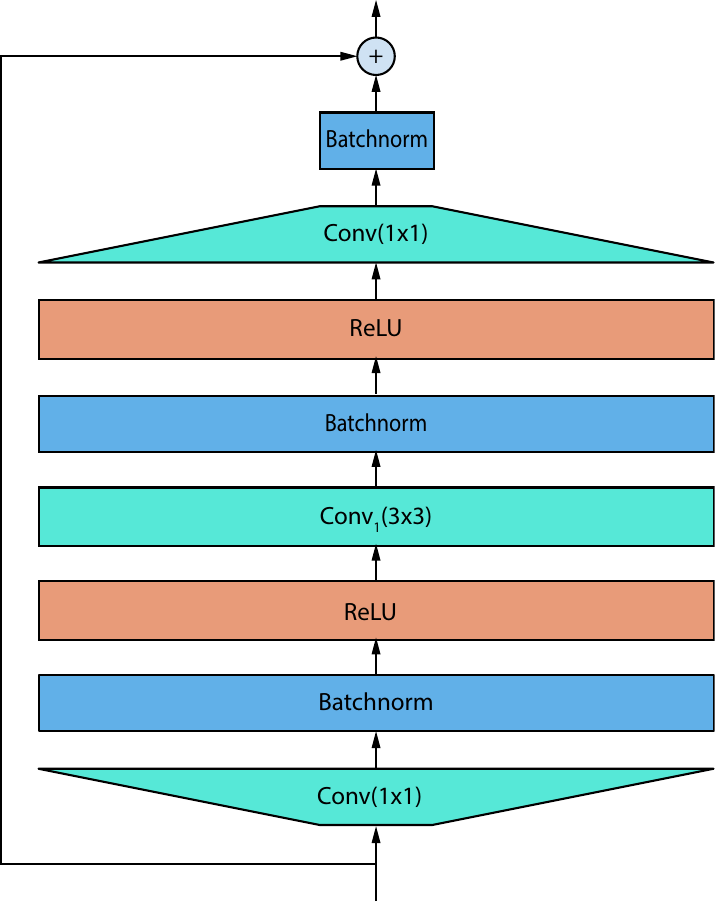}
    \caption{MBConv Inverted Bottleneck}
    \label{f:mbconv-block}
  \end{subfigure}
  \begin{subfigure}[b]{.45\textwidth}
    \includegraphics[scale=0.7,center]{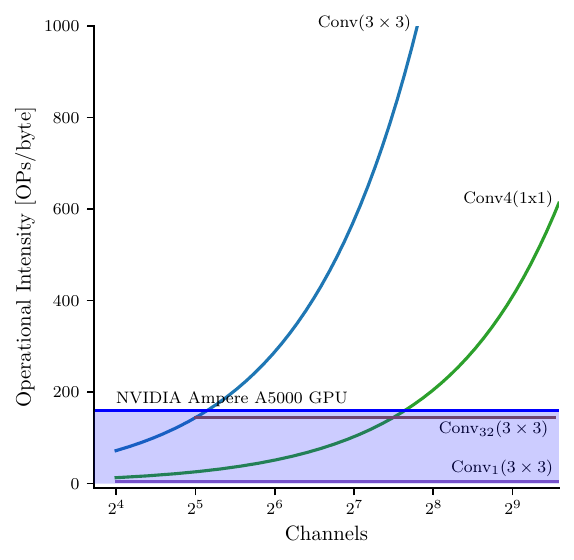}
    \caption{Conv2d Operational Intensity vs. Channels}
    \label{f:waterline-layers}
  \end{subfigure}
  \caption{\textbf{Evolution of convnet blocks and convolutional layers.}
    \textbf{(a).} ResNet34 used \textit{residual blocks} with \lstinline[language=C]!conv2d! layers: Conv($3 \times 3$).
    \textbf{(b).} ResNet50 added \textit{bottleneck} blocks with an expansion ratio equal to four using \textit{point-wise} convolutions: Conv4($1 \times 1$).
    \textbf{(c).} ResNeXT101\_32x4d used grouped-convolutions with group-width equal to 32: Conv\textsubscript{32}($3 \times 3$).
    \textbf{(d).} MobileNetV2 used \textit{inverted residual} blocks with \textit{depth-wise} convolutions: Conv\textsubscript{1}($3 \times 3$).
    \textbf{(e).} Operational intensity of convnet layers as a function of the number of channels. The Conv($3 \times 3$) layers used by early convnets had large operational
    intensity. Point-wise, grouped, and depth-wise convolutions progressively decreased the operational intensity of \lstinline[language=C]!conv2d! layers.
  }
  \label{f:block-evolution}
\end{figure}

Early convnets were composed of stacks of \lstinline[language=C]!conv2d!
layers with kernel size equal to $3 \times 3$ or larger.
\lstinline[language=C]!Conv2d! layers are separated by element-wise
nonlinear activation functions (e.g., ReLU~\cite{hahnloser2000digital, jarrett2009best}, SiLU~\cite{elfwing2018sigmoid, ramachandran2017searching})
 and pooling
 layers that downsampled the feature maps to half resolution~\cite{lecun1998gradient, jarrett2009best}, sometimes merely by
subsampling~\cite{lecun1989backpropagation, he2016deep}. ResNet34 was the zenith of these early convnets~\cite{he2016deep}. The ResNet
\textit{residual block} is shown in Figure \ref{f:resnet-block}.

ResNet50 contributed the \textit{bottleneck block} with \textit{point-wise} convolutions; a degenerate case of the \lstinline[language=C]!conv2d! layer with kernel size equal to $1 \times 1$. Setting $R = S = 1$ simplifies Equation \eqref{f:conv2d} to
\begin{equation}
  Y_{nhwk} = \sum_c^C A_{kc} X_{nhwc} + b
  \label{f:pointwise-conv2d}
\end{equation}
which is just the C-mode (matrix) product (i.e., tensor multiplication~\cite[p.~460]{kolda2009tensor} in the channels dimension) of the weights matrix and input tensor, equivalent to the matrix multiplication
\begin{equation}
  Y_{pk} = \sum_c^C A_{kc} X_{pc} + b
  \label{f:pointwise-matrix-conv2d}
\end{equation}
where dimensions $N \times H \times W$ have been unfolded into a single dimension, $P$. The bottleneck block
is shown in Figure \ref{f:resnet-bottleneck-block}.

ResNeXt~\cite{xie2017aggregated} generalized the \lstinline[language=C]!conv2d! operator
such that the input and output tensors are divided into groups of channels, and weights connect inputs to outputs in the same
group.
This  $\textit{grouped convolution}$ can be written as
\begin{equation}
  Y_{nhwk} = \sum_r^R \sum_s^S \sum_t^T A_{krst} X_{n(h+r)(w+s)(g(k)T+t)} + b
  \label{f:grouped-conv2d}
\end{equation}
where $G$ is the number of groups,  $T = C / G$ is the width of each group, $A \in \R ^{K \times R \times S \times T}$ is the weights tensor, and the group index $g(k) = \floor{\frac{k C}{K T}}$. Grouped convolutions can be understood as \lstinline[language=C]!conv2d! layers with block-diagonal weights tensors. Equation \eqref{f:conv2d} is a special case of Equation \eqref{f:grouped-conv2d} with one group ($G = 1$) and $C$ channels per group ($T = C$).
Figure \ref{f:resnext-bottleneck-block} shows the ResNeXt block.

Xception~\cite{chollet2017xception} and MobileNet~\cite{howard2017mobilenets} used the degenerate case of grouped convolutions where the group width equals a single channel. This \textit{depth-wise convolution} equals
\begin{equation}
  Y_{nhwk} = \sum_r^R \sum_s^S A_{krs} X_{n(h+r)(w+s)\floor{k C / K}} + b
  \label{f:depth-wise-conv2d}
\end{equation}
and in the common case where $K = C$,
\begin{equation}
  Y_{nhwc} = \sum_r^R \sum_s^S A_{crs} X_{n(h+r)(w+s)c} + b
  \label{f:depth-wise-conv2d}
\end{equation}

MobileNetV2~\cite{sandler2018mobilenetv2} used an \textit{inverted residual} block with depth-wise convolution. The MBConv block is pictured in Figure~\ref{f:mbconv-block}.

\subsubsection{Operational Intensity of Conv2d}

The arithmetic complexity of the \lstinline[language=C]!conv2d! layer is
\begin{equation}
n = 2 (N H W K) (C R S) + K
\label{e:conv2d-complexity}
\end{equation}
operations. We can understand this equation as counting $C R S$ multiply-accumulates for every element of the output tensor and K additions for the bias vector.

In the general case of grouped convolution with group-width $T$, Equation \eqref{e:conv2d-complexity} becomes
\begin{equation}
n = 2 (N H W K) (T R S) + K
\label{e:grouped-conv2d-complexity}
\end{equation}
operations, or $T R S$ multiply-accumulates per output.

Assuming the \lstinline[language=C]!conv2d! kernel performs only
\textit{compulsory loads}~\cite{hill1989evaluating}, we reason that it loads
every element of the weights tensor from DRAM exactly once. This assumption is valid for devices with cache that is large enough to
hold the weights tensor of a single layer, which is true for the models and devices considered in this paper.

We also assume that every input element is read from DRAM once and every output
element is written to DRAM. This assumption is true if the on-chip memory is too
small to hold the input and output activation tensors. For now we
consider this to be a simplifying assumption, and it will guide us towards novel
memory-efficient algorithms in the later sections of the paper.

The input tensor has $N H W C$ elements, the output tensor has $N H W K$ elements, the weights tensor has $K C R S$ elements, and the bias vector has $K$ elements. For grouped convolution, the weights tensor
has $K T R S$ elements. Therefore, the general \lstinline[language=C]!conv2d! kernel transfers a total of
\begin{equation}
  b = N H W K + N H W C + K T R S + K
  \label{e:grouped-conv2d-elements}
\end{equation}
elements between the processor and DRAM.

We additionally assume that the input tensor is relatively large with
$N H W \gg T R S$. We accomplish this in practice by setting $N = 128$
for all subsequent experiments. Other papers
that measure vision network latency also used a large batch size~\cite[\dots]{radosavovic2020designing, wightman2021resnet, tan2021efficientnetv2, brock2021high}.

Large batch size is a good fit for the large number of processors found in
discrete, desktop GPUs. It is practical for server-side, batched processing of
images, but not for real-time vision systems. Again, this choice is a
simplifying assumption that we will revisit.

Thus the operational intensity of \lstinline[language=C]!conv2d! with \lstinline[language=C]!float16! arithmetic and two bytes per element equals
\begin{equation}
\begin{split}
\frac{n}{b} &= \frac{(N H W K) (T R S) + K / 2 }{N H W K + N H W C + K T R S + K} \\
&\approx \frac{T R S}{1 + \frac{C}{K} + \frac{T R S}{N H W}} \\
&\approx \frac{T R S}{1 + \frac{C}{K}} \quad \text{if} \quad N H W \gg T R S
\end{split}
\label{e:conv2d-opint}
\end{equation}
where we assumed that the number of input and output elements per channel, $N H W$,
is much greater than the number of operations per output, $T R S$. This also implies
that the activation tensors dominate the DRAM transfers.

Each of the degenerate cases of the \lstinline[language=C]!conv2d! layer reduces
the number of operations and the size of the weights tensor without changing the
size of the input and output tensors. Reduced arithmetic complexity causes the operational intensity to decrease relative to the full \lstinline[language=C]!conv2d! layer.
Figure \ref{f:block-evolution} shows the evolution of convnet blocks and the
operational intensity of the their \lstinline[language=C]!conv2d! layers.

According to Equation \eqref{e:conv2d-opint}, the full \lstinline[language=C]!conv2d! layer
with $T = C$ has operational intensity
\begin{equation}
\frac{n}{b} \approx \frac{C R S}{1 + \frac{C}{K}}
\end{equation}
while point-wise convolution has $R = S = 1$ and $T = C$, yielding
\begin{equation}
\frac{n}{b} \approx \frac{C}{1 + \frac{C}{K}}
\end{equation}
and depth-wise convolution has $T = 1$, giving
\begin{equation}
  \frac{n}{b} \approx \frac{R S}{1 + \frac{C}{K}}
\end{equation}
Thus full \lstinline[language=C]!conv2d! has approximately $R S$ times the operational intensity of point-wise convolution
and $C$ times the operational intensity of depth-wise convolution. Also, it is worth noting that
the operational intensity of grouped and depth-wise convolution is approximately independent of the number of channels, $C$.

As mentioned, we assume that $N H W$ is large in subsequent analyses, which we accomplish by
setting the batch size $N = 128$. In real-time systems that require small batch
size, possibly $N = 1$, the convnet model is typically used as a backbone for an
object detection or semantic segmentation network that usually requires high
resolution images for acceptable accuracy. Still, an input image as small as
$512_H \times 512_W$ will only produce a $16_H \times 16_W$ activation tensor in
the last stage of a typical convnet backbone. So it is worth examining the
possibility $N H W$ is small.

If $N H W$ is approximately equal to $T R S$, then both terms contribute to the operational intensity in Equation~\eqref{e:conv2d-opint}. Conceptually,
the data movement incurred by the weights and input and output activations are all significant. In the extreme, when $T = C$ and  $N H W \ll C R S$, then Equation \eqref{e:conv2d-opint} yields
\begin{equation}
 \frac{n}{b} \approx N H W
 \label{e:conv2d-opint-small-nhw}
\end{equation}
Typically this case would apply when $N H W \ll C$. We can understand this
equation as saying that the cost of loading the weights tensor is amortized over the
pixels in the activation tensor, if that number is small relative to
the width of the network. Real-time systems that require a small batch size in order to
minimize latency might benefit from increasing input image resolution $H \times W$.
They could also use smaller width ($C$) to maintain the total number of operations.

\subsubsection{Degenerate Conv2d and the op:byte Waterline}

Figure \ref{f:waterline-layers} plots the operational intensity of different
types \lstinline[language=C]!conv2d! layers versus the number of input channels, $C$. For
reference, it also shows the op:byte ratio of our baseline processor, the NVIDIA
A5000 GPU. We see that the full \lstinline[language=C]!conv2d! layer with $R = S = 3$ has
relatively large operational intensity even for a small number of channels. Point-wise convolution
with bottleneck expansion ratio equal to four ($\frac{K}{C} = 4$) is ``underwater'' until the
number of channels reaches a relatively large number. Grouped convolution with $T = 32$, as used by
ResNeXT101\_32x4d, has a flat operational intensity that is underwater. Depth-wise convolution with $R = S = 3$ has operational intensity
close to zero.

Because on-chip SRAM has greater bandwidth than off-chip DRAM, hardware designers can
improve the computational efficiency of memory-bound kernels by increasing
the SRAM size until all the activations fit on-chip. However, SRAM size
contributes to the die size which determines cost.
A kernel that uses a larger workspace requires
a greater die size to achieve the same arithmetic throughput. It achieves its
performance at a greater cost than a kernel that uses memory efficiently.

\FloatBarrier

\subsection{Waterline Analysis of Convnet Models}
\label{s:waterline-analysis-baseline}

In this section we use Equation \eqref{e:total-attainable-latency} to investigate the minimum attainable latency of different convnet models.

\begin{figure}[h]
  \includegraphics[width=\textwidth]{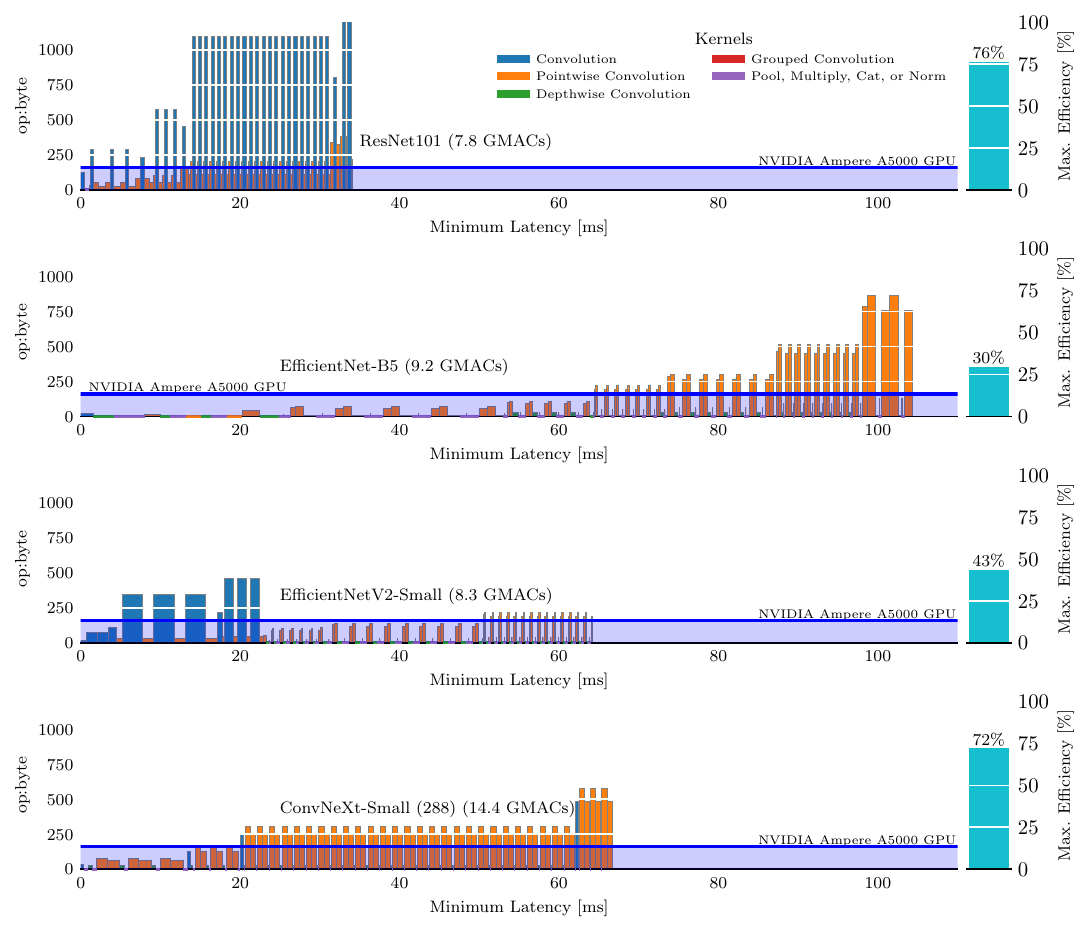}
  \caption{\textbf{Waterline analysis} of baseline models. These plots compare
    the operational intensity of individual layers with the op:byte
    waterline of the NVIDIA Ampere A5000 GPU. Single-layer kernels are
    often memory bound because their low operational intensity is
    ``underwater.'' Memory bound kernels have higher attainable latency and lesser attainable computational efficiency (max
    efficiency). Again, we used \lstinline[language=C]!float16! arithmetic and batch size equal to 128.}
  \label{f:waterline-baseline-models}
\end{figure}

Figure \ref{f:waterline-baseline-models} shows the minimum attainable latency and the corresponding maximum computational efficiency for baseline
convnet models. We plot
the operational intensity of each kernel on the y-axis and the corresponding
minimum attainable latency on the x-axis. Kernels that rise above the
processor's op:byte \textit{waterline} are compute bound, others are memory
bound. We also compute the maximum computational efficiency for each network (with Equation \eqref{e:max-comp-eff}) and show it next to the waterline plot.

We assume layer-by-layer execution of the network, while allowing for trivial operator
fusions that combine convolution, bias, residual shortcut addition, and
activation functions in a single kernel. These are the simple kernel fusions
implemented in PyTorch Inductor and other available inference engines.

Figure \ref{f:waterline-baseline-models} predicts relatively high maximum efficiency for ResNet101 and ConvNeXt-Small, which agrees with the popular sentiment that those models are computationally efficient in deep learning frameworks.

ResNet101 has many \lstinline[language=C]!conv2d! layers with large operational intensity. It also has many point-wise convolution layers that vary from low operational intensity in the early stages of the network to moderate operational intensity in the late stages where the number of channels is greater.

ConvNeXt-Small has many point-wise convolution layers with large operational intensity due to the fact they use a large number of channels. It also has depth-wise convolution layers with extremely low operational intensity, but these transfer relatively little data, because they are placed at the block's bottleneck where the activations tensor has fewer channels.

EfficientNet-B5 has relatively low maximum attainable efficiency because of the small number of channels in the early stages of the network. Also, it has depth-wise convolution layers with low operational intensity, and these layers operate on the hidden-layer tensors in the bottleneck blocks where the number of channels is greatest.

These results suggest that narrow (small number of channels per layer) models are computationally inefficient. This observation agrees with the work of Dollar et al.~\cite{dollar2021fast} who saw that computational efficiency decreases as the number of network activations increases. They reasoned that models should have more channels and fewer layers to increase the ratio of activations to operations. We understand this as a simple strategy to increase the operational intensity of point-wise convolutions.

But this creates a dilemma, because the narrower and deeper models like EfficientNet have greater model efficiency. Ideally one could improve computational efficiency without sacrificing model efficiency.

\FloatBarrier

\subsection{Mediant Operational Intensity}
\label{s:mediant-opint}

Several papers applied the roofline performance model directly to an entire neural network~\cite[\dots]{li2021searching, jouppi2021ten, jouppi2017datacenter}. These papers computed the operational intensity of a network
simply by dividing the total operations by the total bytes transferred to and from DRAM,
\begin{equation}
  \text{mediant operational intensity} = \frac{\sum_i{n_i}}{\sum_i{b_i}}
\label{e:mediant-opint}
\end{equation}
for layers $i$, with $n_i$ operations and $b_i$ bytes transferred per layer. Mathematicians recognize Equation~\eqref{e:mediant-opint} as the \textit{mediant} of the fractions $\frac{n_i}{b_i}$. The term originates from the inequality
$\frac{a}{b} < \frac{a + c}{b + d} < \frac{c}{d}$~\cite{guthery2011motif, fowler1991approximation}.

If we were to interpret Equation~\eqref{e:mediant-opint} as the operational intensity of a set of kernels, then
we would infer that the kernels run concurrently. The roofline model requires this assumption,
because the latency of computation cannot hide the latency of data movement unless both run concurrently.

Otherwise, if the kernels run sequentially, then Equation~\eqref{e:mediant-opint} only approximates
the waterline performance model presented earlier. It is correct when the kernels are all memory-bound or all compute-bound.
But it overestimates the attainable performance of a
sequence that has both memory-bound and compute-bound kernels. Because Equation~\eqref{e:mediant-opint} adds the
excess operations of the compute-bound kernels to the operations of the memory-bound kernels,
the sequence appears to be less memory-bound than it truly is.

\begin{figure}[h]
\begin{subfigure}[b]{0.45\textwidth}
  \includegraphics[scale=1.0,center]{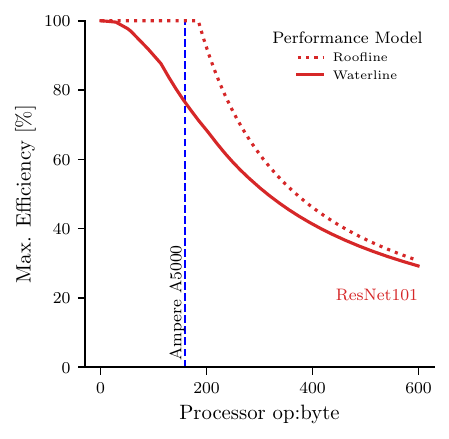}
  \label{f:average-opint-resnet}
\end{subfigure}
\begin{subfigure}[b]{0.45\textwidth}
  \includegraphics[scale=1.0,center]{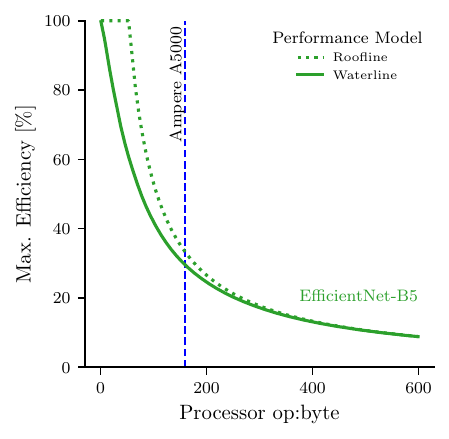}
  \label{f:average-opint-efficientnet}
\end{subfigure}
\\
\begin{subfigure}[b]{0.45\textwidth}
  \includegraphics[scale=1.0,center]{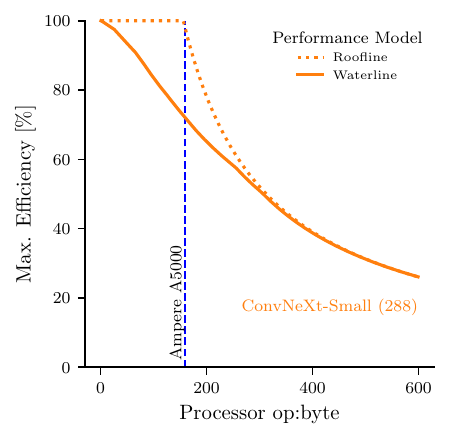}
  \label{f:average-opint-convnext}
\end{subfigure}
\begin{subfigure}[b]{0.45\textwidth}
  \includegraphics[scale=1.0,center]{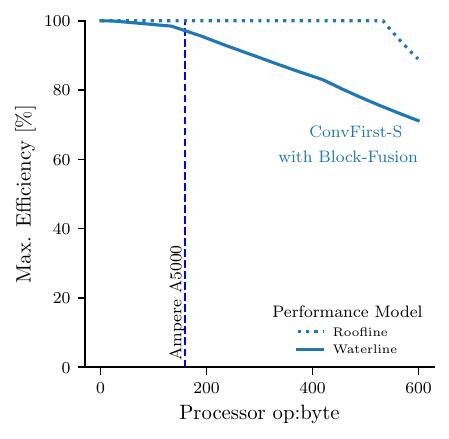}
  \label{f:average-opint-convfirst}
\end{subfigure}
\caption{Comparison of the attainable computational efficiency
  calculated by the waterline and roofline performance models.
  Roofline was originally intended as a performance model for a single parallel kernel~\cite{williams2009roofline}. Hence roofline overestimates the attainable efficiency for a sequence of kernels, if some of the kernels are compute bound
  and others are memory bound. Waterline is accurate regardless, because it measures how each kernel contributes to the minimum latency of the sequence. ConvFirst is our new model. See Section~\ref{s:ConFirstNet} for details.
}
\label{f:average-opint}
\end{figure}

How accurate is Equation~\eqref{e:mediant-opint} when applied to a convnet?
Figure~\ref{f:average-opint} plots the maximum attainable efficiency of different
convnets versus the processor op:byte ratio. It shows both the roofline and waterline performance models. Again, we use batch size
equal to 128 and 2 bytes per tensor element. For processors with very high
op:byte ratios, all layers of the convnets are memory bound, and the roofline
model (using the mediant operational intensity) agrees with the waterline model.
For low to moderate op:byte ratios, the roofline model evaluates the convnets as
compute bound, while the waterline model recognizes that some of the layers are
memory bound. This discrepancy causes the roofline model to overestimate the
attainable efficiency.

For models like ResNet101 and ConvNeXt-Small, which have some layers with
relatively high operational intensity, roofline drastically overestimates the
attainable efficiency, unless the processor has a very large op:byte ratio. For
models like EfficientNet-B5, which have fewer layers with large operational intensity, roofline is closer to the
waterline model.

\FloatBarrier

\section{Block Fusion: An Optimization for Degenerate Conv2d Layers}\label{s:block-fusion}

In this section, we improve the operational intensity of inverted residual blocks
by fusing their layers into a single kernel. We find a path through each block's
computational graph that creates temporal locality and avoids DRAM memory
transfers. The inverted residual block becomes a module with low-bandwidth
interface and computationally dense implementation. We call this technique
\textit{block fusion}.

Block fusion is faster than layer-by-layer execution, because the degenerate
\lstinline[language=C]!conv2d! layers used in modern convnets have low
operational intensity when computed separately.

To navigate the complex space of kernel designs, we introduce a
simple computational model called the ``tensor machine.''

\subsection{Tensor Machines}

We believe that the complexity of low-level source code obscures the high-level
algorithmic ideas that are essential for efficient kernels. We devised the
tensor machine as a picture language for drawing simple kernel diagrams that
capture the essence of our efficient algorithms.

\begin{figure}[h]
\includegraphics[scale=0.75]{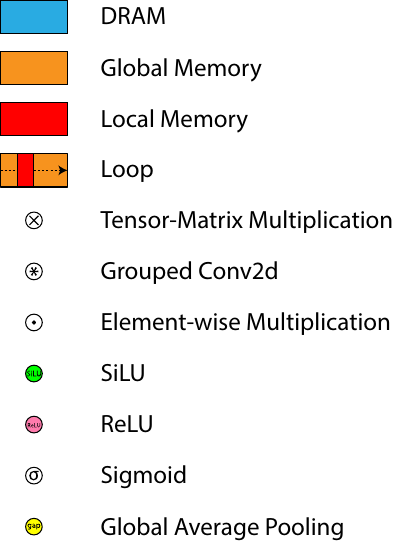}
\caption{\textbf{The tensor machine} is a simple abstract computer for
  designing convnet kernels. It has three memory spaces: DRAM (slow, off-chip
  memory), global memory (on-chip memory shared by all processors), and local
  memory (fast memory internal to each processor). It uses basic high-level
  operations, including matrix multiplication (or channels-mode matrix product), grouped conv2d,
  element-wise multiplication, global average
  pooling, and activation functions. The \textit{loop} symbol indicates a
  for-loop that iterates over memory tiles, loading one tile at a time into a
  local memory buffer. }
\label{f:tensor-machine-legend}
\end{figure}

The \textit{tensor machine} is a simple abstract computer that helps us visualize
kernels and analyze their theoretical efficiency. It uses simple high-level
operations that act on tensors and places each tensor in DRAM, global memory, or
local memory. Tensor machines define a kernel by showing its
operations and data movement using the symbols in Figure
\ref{f:tensor-machine-legend}.

Tensor machine operations include matrix multiplication (or channels-mode product of an activation tensor and a weights matrix~\cite[p.~460]{kolda2009tensor}), grouped conv2d, element-wise multiplication, activation
functions, and global average pooling.

The tensor machine's three memory regions are DRAM, global memory, and local
memory. DRAM is slow, off-chip memory with large capacity. Global memory is on-chip memory
that is accessible by all processors. Local memory is fast on-chip memory
internal to each processor.

Global memory functions as a cache for weights and
overlapping inputs and as a shared memory for inter-processor communication.

Global and local memory have limited capacity. In this paper we merely assume
that global memory is sufficiently large to hold the weights of a single
residual block and a small number of activations. This assumption is reasonable
for all the models in this paper.

Local memory is only large enough to store small tiles of activations and
weights tensors. High-resolution tiles reduce the global memory bandwidth requirement, because more computation is
performed for each weight value that is transferred from the global memory cache. In this
paper, we defer analysis of the exact local memory capacity and tile sizes to
the platform-specific kernel development stage.

We use waterline analysis to compute the minimum attainable latency for a kernel
and its tensor machine with a given op:byte ratio. We only count the
multiply-accumulate operations of tensor-matrix multiplication and conv2d
operators, all other operators do not contribute to the operational
intensity. This approximation is useful for neural networks because
tensor-matrix multiplications and conv2d operations dominate their arithmetic
complexity.

This simple machine facilitates the high-level choices that dominate performance and avoids the low-level
details that create complexity.

\subsection{Feed-Forward Network}

Position-wise Feed-Forward Network (FFN) is a block used prominently by Transformers~\cite{vaswani2017attention} and ConvNeXt~\cite{liu2022convnet}. It is also called Multi-Layer Perceptron (MLP)~\cite{tolstikhin2021mlp}.
FFN performs two consecutive linear transformations with an element-wise activation function after the first:
\begin{equation}
\ffn(X) = \phi(X U + a) V + b
\label{e:ffn}
\end{equation}
where $X \in \R^{P \times C}$ is the input, $U \in \R^{C \times \alpha C}$ and $V \in \R^{\alpha C \times C}$ are the weights, $a \in \R^{\alpha C}$ and $b \in R^C$ are the bias, and $\phi$ is an activation function, with $P$ as the number of samples, $C$ as the number of
channels, and $\alpha$ as the expansion factor.

The most direct algorithm to compute $\ffn$ operates layer-by-layer, first computing hidden layer $Y \in \R^{P \times \alpha C}$ as
\begin{align*}
Y = \phi(X U + a)
\end{align*}
then computing output $Z \in \R^{P \times C}$ as
\begin{align*}
Z = Y V + b
\end{align*}

The layer-by-layer algorithm creates a large hidden layer $Y$ with $\alpha$ times as many channels as the input and output layers $X$ and $Z$. In a stack of FFN blocks, reading and writing $Y$ dominates the memory cost of the layer-by-layer algorithm, accounting for $\frac{\alpha}{\alpha+1}$ of all activation memory.

We will derive a memory-efficient algorithm that practically eliminates the hidden layer activations. This algorithm is a special case of the one that was proposed by Sandler et al. for the MBConv block~\cite{sandler2018mobilenetv2}.

First, compute a single channel $Y^r$ of the hidden layer by multiplying the input by column $U^r$ of weights
\begin{equation}
  Y^r = \phi(X U^r + a_r)
  \label{e:ffn-yr}
\end{equation}
and exploiting the fact that $\phi$ is an element-wise function.\footnote{For matrix $A$, we write $A_i$ for row $i$ and $A^j$ for column $j$~\cite[p.54]{apostolv2}.}

Second, multiply channel $Y^r$ by row $V_r$ to form an outer product, and the sum of the products over $r$ equals output matrix $Z$:

\begin{align*}
Z = \sum_{r=1}^{\alpha C} Y^r V_r + b
\end{align*}

Substituting Equation \eqref{e:ffn-yr} for $Y^r$ yields the complete formula:
\begin{equation}
  \ffn(X) = \sum_{r=1}^{\alpha C} \phi(X U^r + a_r) V_r + b
  \label{e:ffn-eff}
\end{equation}

Equation \eqref{e:ffn-eff} defines a memory-efficient algorithm that computes a single channel of the hidden layer, immediately uses it to update the output layer, and then discards it. The algorithm iterates over all hidden channels and reuses the same small workspace to store each of them.
Therefore, only a small fraction of the hidden layer exists in memory at any given time.

From an algorithmic standpoint, Equation \eqref{e:ffn-eff} implements a
\textit{loop fusion} optimization~\cite[p.~254]{kennedy2001optimizing}. In the
layer-by-layer algorithm, the first linear transformation loops over $r \in [1~\ldotp\ldotp~\alpha C]$,
producing hidden channels $Y^r$. The second linear transformation
also loops over $r$, consuming $Y^r$. The large number of hidden channels $\alpha C$
causes the time elapsed between the production and consumption of $Y^r$ to be large. Therefore, a large workspace is needed to
store the full tensor $Y$. Loop fusion increases temporal locality by producing and consuming
each $Y^r$ in the same loop iteration. Both the size of the workspace and
the communication cost of the algorithm are reduced.

From a model architecture standpoint, Equation~\eqref{e:ffn-eff} virtually eliminates
hidden layer $Y \in \R^{P \times \alpha C}$. The $\ffn$ block becomes a single layer
with weights $U \in \R^{C \times \alpha C}$ and $V \in \R^{\alpha C \times C}$, input $X \in \R^{P \times C}$, output $Z \in \R^{P \times C}$, no hidden layer, and large
operational intensity.

\FloatBarrier

\subsection{ConvFirst Block}

\begin{figure}[h]
  \begin{subfigure}[b]{.3\textwidth}
    \label{sf:fusedmbconv}
    \includegraphics[scale=0.35,center]{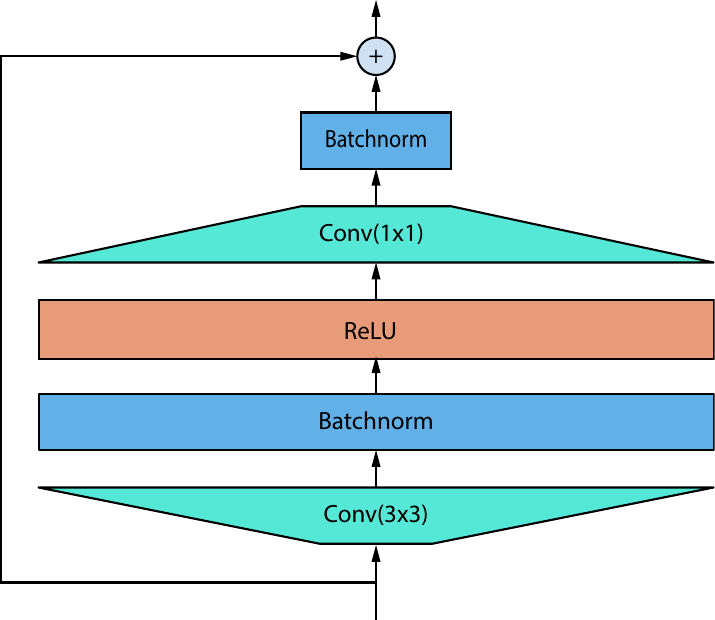}
    \caption{FusedMBConv}
  \end{subfigure}
  \begin{subfigure}[b]{.3\textwidth}
    \label{sf:convfirst-stride1}
    \includegraphics[scale=0.35,center]{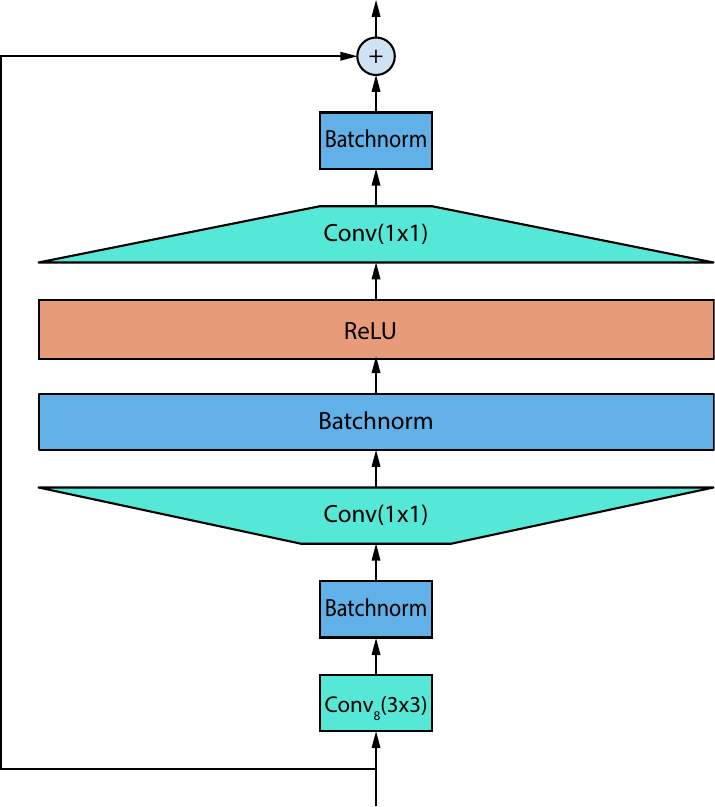}
    \caption{ConvFirst}
  \end{subfigure}
  \begin{subfigure}[b]{.3\textwidth}
    \label{sf:convfirst-stride2}
    \includegraphics[scale=0.35,center]{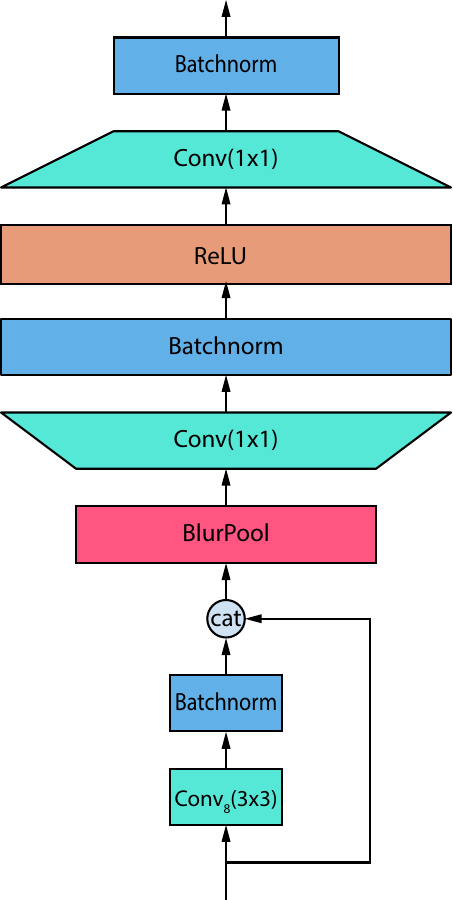}
    \caption{ConvFirst Stride 2}
  \end{subfigure}
  \caption{
    \textbf{(a).} The FusedMBConv block, as used in EfficientNet-EdgeTPU~\cite{efficientnet_edgetpu} and EfficientNetV2~\cite{tan2021efficientnetv2}, has a \lstinline[language=C]!conv2d!$(3 \times 3)$ layer that uses a large number of arithmetic operations. The block also affords a memory-efficient block-fusion implementation~\cite{lavin2021pie}. \textbf{(b).} We created the ConvFirst block to be a light-weight mimic of FusedMBConv. ConvFirst also affords block fusion but uses far fewer operations. ConvFirst is very similar to ConvNeXt~\cite{liu2022convnet}. \textbf{(c).} ConvFirst uses BlurPool~\cite{zhang2019making} for downsampling to improve shift invariance of downstream tasks.
  }
  \label{f:convfirst}
\end{figure}

We designed a convnet block with the goal of exploiting the FFN algorithm from Equation
\eqref{e:ffn-eff}. The first layer of the block is a grouped convolution. We
chose group width equal to 8 channels because it matches the size of the matrix multiply
implemented by the NVIDIA tensor core instruction \lstinline[language=C]!mma.m16n8k8.f16!~\cite{nvidia_ptx}. An
FFN layer follows the grouped convolution. A residual shortcut connects the
block input to the output, and a batchnorm layer~\cite{ioffe2015batch} follows each linear layer. The downsampling
variant of ConvFirst used BlurPool~\cite{zhang2019making}. Figure \ref{f:convfirst} shows the block diagrams.

We call the block \textit{ConvFirst} because it puts the convolutional layer
before the FFN layers. ConvFirst differs from ResNet and MBConv, which place
the convolution in the middle of the block.

Putting the convolution first means that there is no overlap in the receptive
fields of the rest of the block's layers. FFN layers have a
one-to-one dependency between input and output pixels. Therefore, we can divide
the output tensor into many tiles and compute each of them in parallel
without any inter-tile communication or over-compute. Only the input tiles
overlap, and the overlap is handled gracefully by the memory cache, so each input value is
loaded from DRAM at most once.

ConvFirst uses the ReLU~\cite{hahnloser2000digital, jarrett2009best} activation function because it can be computed
efficiently. The cost of the computation is amortized over
the block's input channels, so an advanced function like SiLU~\cite{elfwing2018sigmoid, ramachandran2017searching} would be
relatively expensive if the number of channels were small. ReLU does cause a slight
accuracy drop relative to SiLU in our model experiments, and SiLU computation is
reasonably efficient on processors that accelerate the
\lstinline[language=C]!tanh! special function. Therefore, SiLU might be preferable in
larger models.

ConvFirst is very similar to ConvNeXt, which was discovered
independently~\cite{liu2022convnet}. ConvFirst and ConvNeXt are basically the same block
with different normalization, group-width, and kernel-size.

\FloatBarrier

\subsubsection{Block-Fusion Kernel for ConvFirst}

We designed a tensor machine kernel that implements the ConvFirst block. The
\textit{block-fusion} strategy casts the entire block as a single computation
with three weights tensors and no hidden activations. Actually each hidden
activation exists for a moment in local memory, but it is invisible
to the global memory system. Therefore, ConvFirst blocks have low
communication cost and large operational intensity, even when the number of channels is small.

\begin{figure}[h]
  \begin{subfigure}{\textwidth}
    \caption{ConvFirst with layer-by-layer execution.}
    \centering
    \label{sf:convfirst-layerwise}
    \includegraphics[scale=0.65]{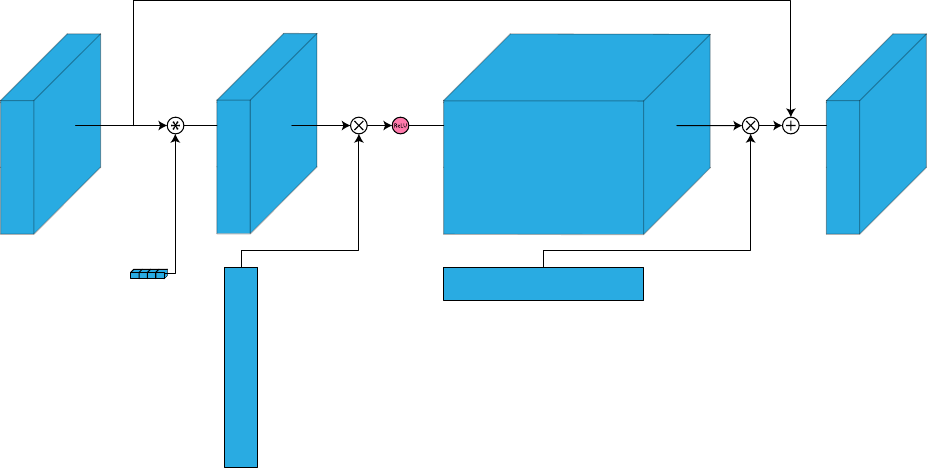}
  \end{subfigure}
  \par\bigskip
  \begin{subfigure}{\textwidth}
    \caption{ConvFirst with block-fusion.}
    \centering
    \label{sf:convfirst-block-fusion}
    \includegraphics[scale=0.65]{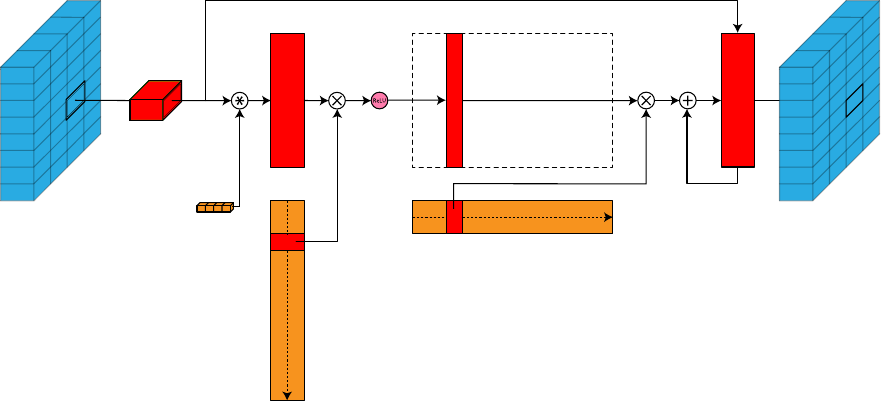}
  \end{subfigure}
  \par\bigskip
  \begin{subfigure}{\textwidth}
    \caption{ConvFirst with block-fusion and scaling.}
    \centering
    \label{sf:convfirst-block-fusion-scaling}
    \includegraphics[scale=0.65]{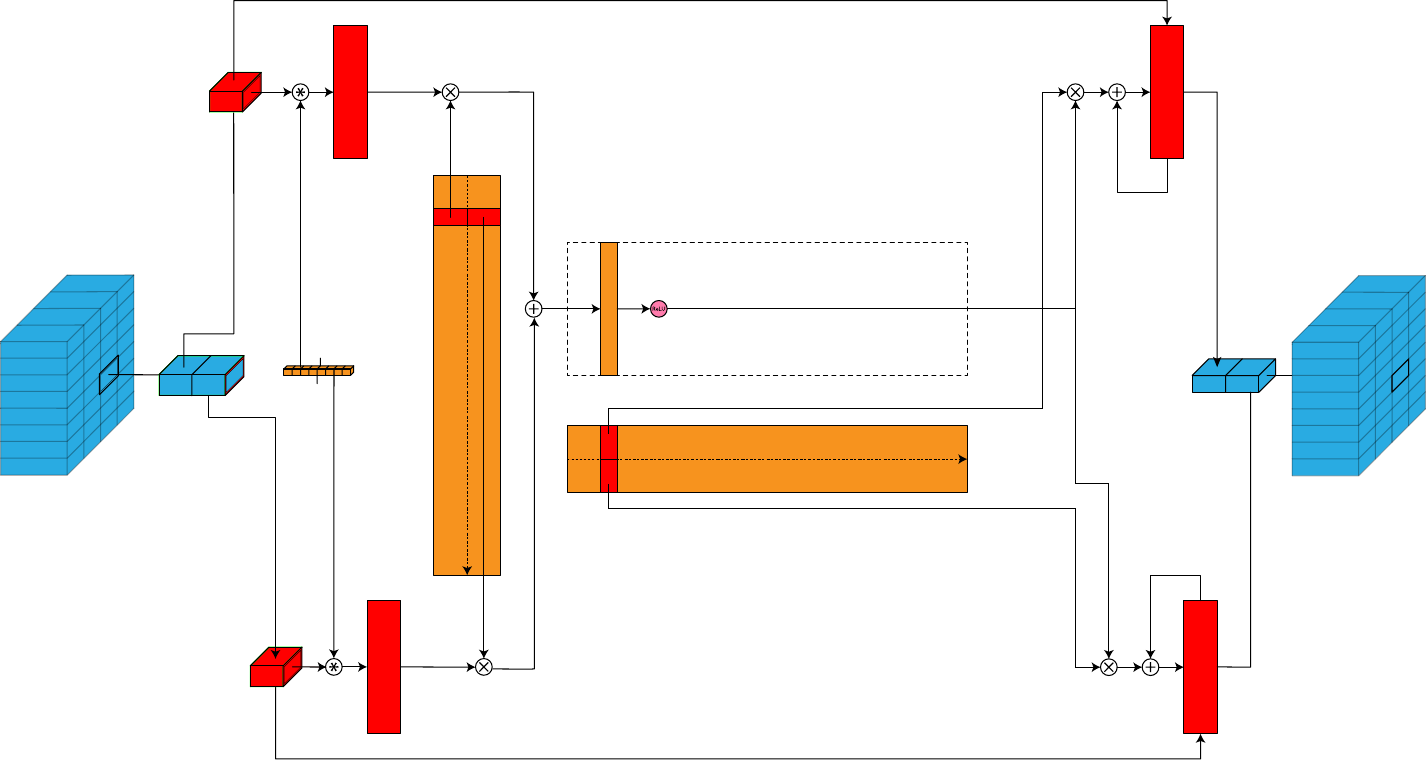}
  \end{subfigure}
  \caption{
    \textbf{Tensor machine} for ConvFirst kernels.
    \textbf{(a).} Layer-by-layer execution creates a large tensor in DRAM for the hidden-layer activations.
    \textbf{(b).} Block-fusion computes all layers simultaneously using small tiles in local memory (red).
    \textbf{(c).} Block-fusion scales to support a large number of channels by partitioning the input and output tiles channel-wise across multiple processors.
    For large $N H W$, the weights matrices are almost always in global memory (orange) cache.
  }
  \label{f:convfirst-tensor-machine}
\end{figure}

Figure \ref{sf:convfirst-layerwise} shows the tensor machine that executes ConvFirst layer-by-layer. This execution plan produces a large DRAM tensor for
the hidden-layer activations.

Figure~\ref{sf:convfirst-block-fusion} shows our block-fusion kernel. It loads an
input tile from DRAM into local memory. It computes the grouped-convolution of
the input tile and retains the result in local memory. It also initializes the
output accumulators with the input-tile values, thereby implementing the residual
shortcut connection.

Next it computes the FFN layers using the loop fusion algorithm defined in
Equation~\eqref{e:ffn-eff}. A single loop over hidden-layer channels produces
activations with the first linear transform and
projects them onto the output tile with the second. After the last loop iteration, the ConvFirst block is complete and the
output tile is stored to DRAM.

Each loop iteration processes a small number of hidden-layer channels,
and a small buffer holds the working set in local memory. Gone is
the large DRAM buffer that layer-by-layer execution uses.

The block-fusion kernel requires that the number of channels is small enough
that the input and output tiles fit in local memory. This requirement is
satisfied for the small vision models we create in the later part of this paper and the
NVIDIA GPU on which we implement the kernel. It would not be true for large models.

The kernel variant shown in Figure\ref{sf:convfirst-block-fusion-scaling} scales
the block-fusion kernel for a large number of channels. It
partitions the input and output tiles channel-wise across two processors. Each
processor loads half the input channels and produces half the output channels. A
small tensor in global memory accumulates the partial sums produced by each
processor to form the hidden-layer channels.

The cost of the add operation in global memory is amortized over the channels of the
input and output tensors, so we expect it to be insignificant when the number of
channels is large. For a very large number of channels, we would partition them
across four or more processors.

Batchnorm layers are constant affine transforms during inference, so 
we fold them into the weights and bias of the preceding
\lstinline[language=C]!conv2d! layer.

\FloatBarrier

\subsubsection{Waterline Analysis of ConvFirst}

\begin{figure}[h]
\includegraphics[width=\textwidth]{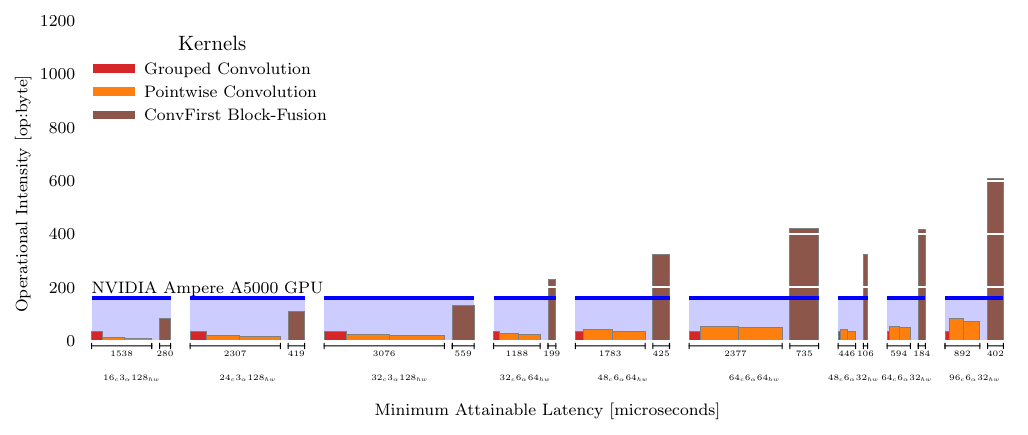}
\caption{\textbf{Waterline plot} for the ConvFirst block with single-layer and block-fusion kernels. The vertical axis shows the operational intensity of the kernels, and the horizontal axis shows the minimum latencies derived from roofline analysis. A horizontal blue line indicates the op:byte ratio (i.e., the \textit{waterline}) of the A5000 GPU. ``Underwater'' kernels are memory bound. The block-fusion kernels avoid writing the activations of the hidden layer to DRAM. Therefore, they have significantly greater operational intensity and can attain lower latency.}
\label{sf:convfirst-waterline}
\end{figure}

Figure \ref{sf:convfirst-waterline} shows the waterline analysis of the
ConvFirst kernels for a range of block parameters with the op:byte ratio of
the A5000 GPU as the waterline. The plot compares the operational intensity and
minimum attainable latency for the single-layer kernels and the block-fusion
kernel. Each ``well'' shows the results for a different choice of block
parameters.

Single-layer kernels are memory bound for all blocks, while the
block-fusion kernels are only memory bound when the expansion ratio ($\alpha$) is
small. When the expansion ratio equals 6, the block-fusion kernels are compute bound
for as few as $32$ channels. The minimum attainable latency of the
block-fusion kernels is significantly lower for all blocks. The speedup is
greatest when the number of channels and expansion ratio is smallest, where the
point-wise convolutions are severely memory bound.

\FloatBarrier

\subsection{MBConv Block}

The MBConv block was originally developed for MobileNetV2 as a memory-efficient
block suitable for convnets that run on small mobile
devices~\cite{sandler2018mobilenetv2}. EfficientNet added Squeeze \& Excitation
layers to the MBConv block and scaled the model to very large size
and accuracy while achieving state-of-the-art model and parameter
efficiency~\cite{tan2019efficientnet}.

\begin{figure}[h]
  \begin{subfigure}[b]{.45\textwidth}
    \label{sf:mbconv-stride1}
    \includegraphics[scale=0.5,center]{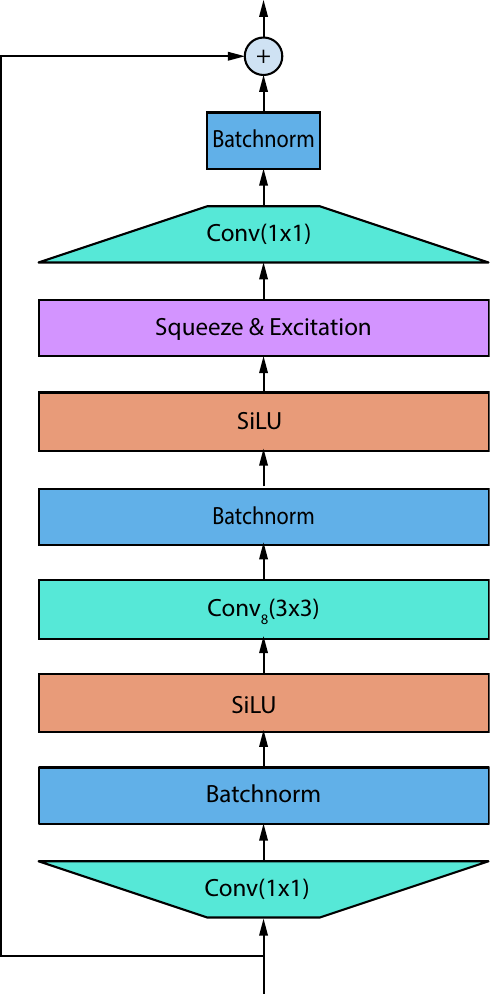}
    \caption{MBConv}
  \end{subfigure}
  \begin{subfigure}[b]{.45\textwidth}
    \label{sf:mbconv-stride2}
    \includegraphics[scale=0.5,center]{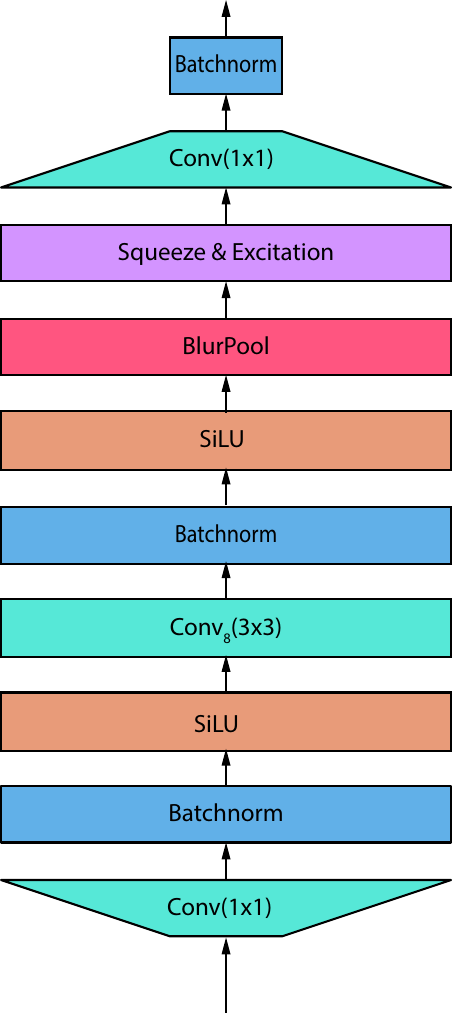}
    \caption{MBConv stride 2}
  \end{subfigure}
  \caption{\textbf{MBConv} block with Squeeze \& Excitation. We modified the traditional MBConv block to use group-width 8 convolution because that operation maps perfectly to NVIDIA tensor cores. We added a BlurPool~\cite{zhang2019making} layer after the grouped convolution layer to implement downsampling.}
  \label{f:mbconv}
\end{figure}

Figure \ref{f:mbconv} shows our variant of MBConv. We substituted grouped
convolution for the traditional depth-wise convolution, to achieve
better utilization on matrix multiply accelerators. On NVIDIA GPUs with \lstinline[language=C]!float16!
tensor core arithmetic, we used group-width equal to eight channels.

Our MBConv variant uses BlurPool for downsampling~\cite{zhang2019making}. The purpose
of BlurPool is to improve shift invariance in downstream tasks, like object
detection and semantic segmentation. In our model experiments, we used BlurPool
with the Triangle-3 filter.

The waterline analysis of EfficientNet in Section
\ref{s:waterline-analysis-baseline} showed that all layers of the MBConv block
are memory bound in the early stages of the network, where the number of
channels is small. Therefore, we are interested in MBConv for the late
stages of the network, where the number of channels is large and the resolution
(height and width) of the activation tensor is small. We designed an MBConv
block-fusion kernel that targets this hyperparameter domain.

\FloatBarrier

\subsubsection{Block-Fusion Kernel for MBConv}
\label{s:mbconv-tensor-machine}
\begin{figure}[h]
  \centering
  \includegraphics[width=\textwidth]{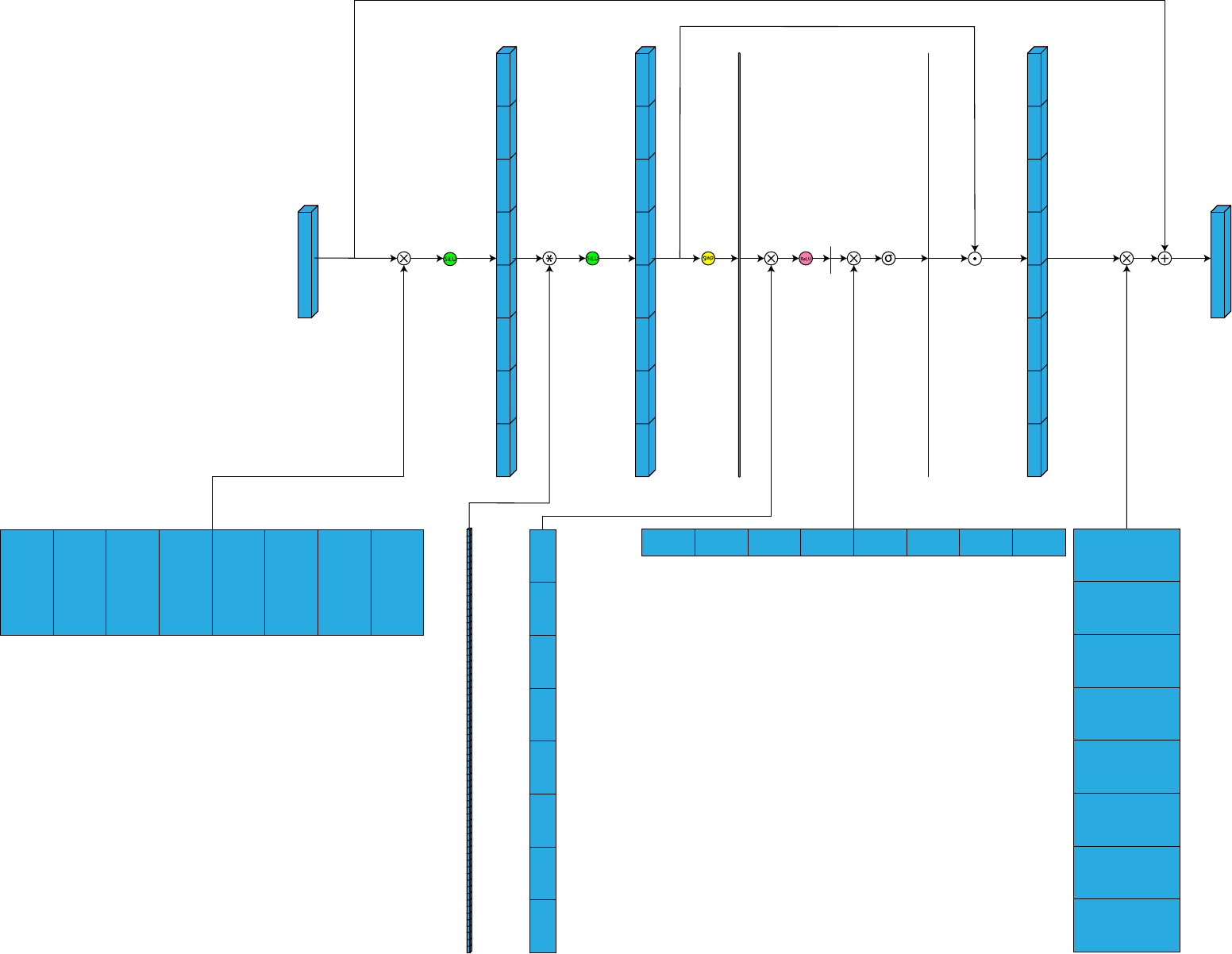}
  \caption{\textbf{Tensor machine} for the layer-by-layer execution of MBConv + Squeeze \& Excitation. When used in the late stages of a convnet, MBConv blocks have a relatively large number of channels and low-resolution feature maps. The large number of channels causes relatively large operational intensity for the point-wise convolution layers. But the grouped-convolution and SE layers have very low operational intensity, because they perform few operations per activation. We show the weights matrices and activation tensors divided into eight partitions that correspond to the matrix and tensor blocks used by the block-fusion kernel in the next figure.}
  \label{f:mbconv-tensor-machine}
\end{figure}

Figure \ref{f:mbconv-tensor-machine} shows the tensor machine for
layer-by-layer execution for the MBConv block with Squeeze \& Excitation. The tensor
shapes are representative of those found in the later stages of a convnet, where
the resolution of the activation tensor is low and the number of channels is large.
These hyperparameters produce small activation tensors and relatively large
weights matrices. For the model sizes considered in this paper, the
weights tensors fit in global memory of a contemporary GPU.

We show the weights matrices and activation tensors partitioned into eight blocks
each, which correspond with the blocks used by each processor in the
block-fusion algorithm, described below.

\begin{figure}[h]
  \includegraphics[width=\textwidth]{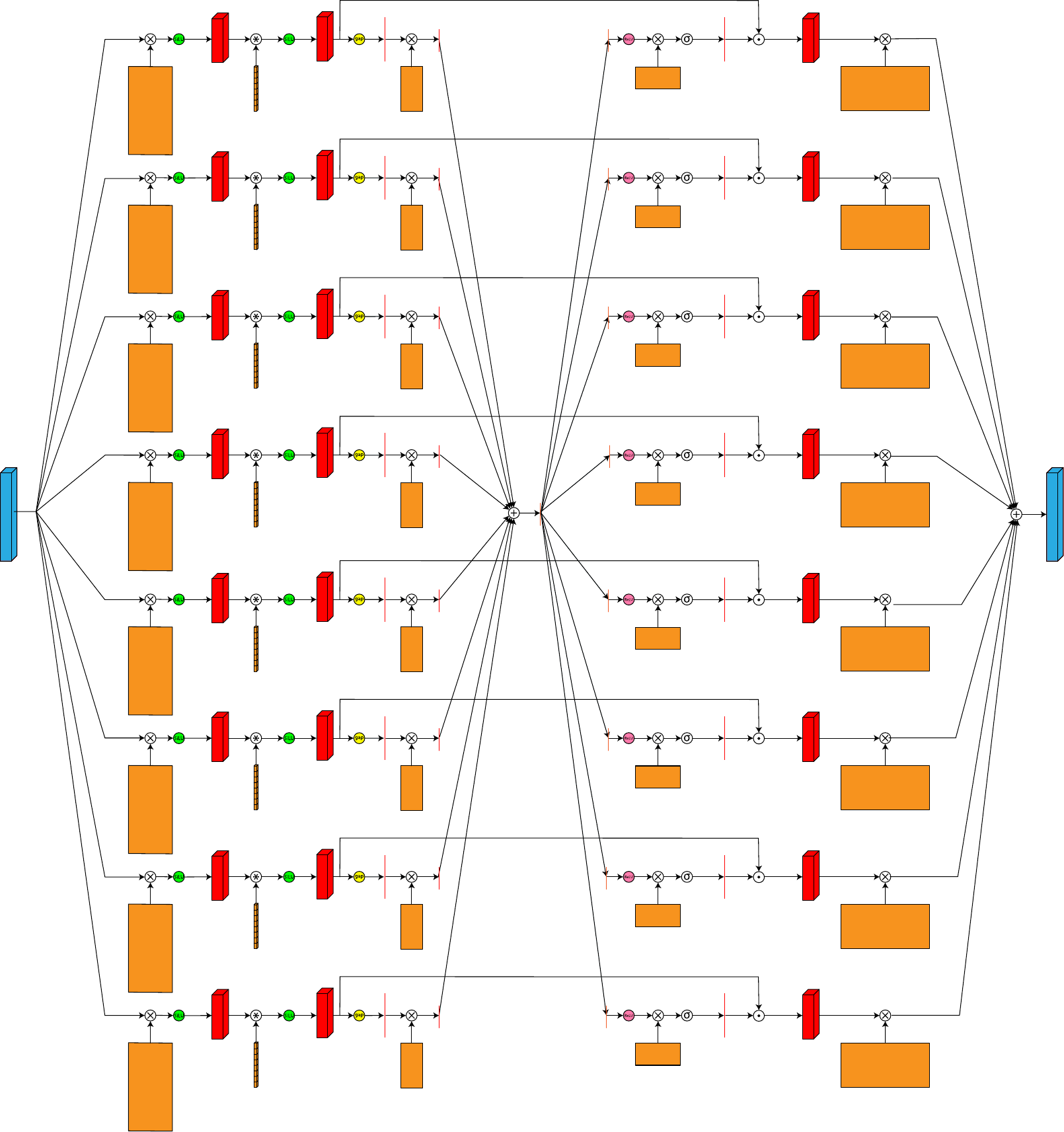}
  \caption{\textbf{Tensor machine} for the block-fusion of MBConv + Squeeze \& Excitation. Multiple processors compute the MBConv block in parallel, with each processor computing a partition of the hidden-layer channels. The hidden-layer activations never leave the local memory of the processor that computes them. Each processor loads only the weights corresponding to its channels assignment.  The ``squeeze'' operation of the SE layer accumulates the contributions from each processor onto a tensor in global memory. The residual shortcut is implemented by accumulating the output tensor onto the input tensor. The kernel has great operational intensity but uses a lot of global memory bandwidth. When implemented on an NVIDIA GA102~\cite{nvidia2022ga102}, the performance is limited by the latency of inter-processor synchronization. The kernel would be more efficient on the Hopper GPU architecture~\cite{nvidia2022h100}, using distributed shared memory for inter-processor communication.}
  \label{f:mbconv-block-fusion-tensor-machine}
\end{figure}

Figure \ref{f:mbconv-block-fusion-tensor-machine} shows the tensor machine
for a block-fusion kernel. We documented this same kernel in a previous article~\cite{lavin2023mbconv}.

The block-fusion kernel uses multiple processors, assigning a group of hidden-layer
channels to each. Each processor computes its own group of hidden-layer channels using the
corresponding block of weights from the first point-wise convolution and storing the activations in local memory. Then it
applies the SiLU activation function.

Then each processor computes the grouped convolution on its channels, again applying the SiLU activation function, producing the convolution
layer output, which is also stored in local memory. If the width of the grouped convolution evenly divides the number of channels assigned to each processor, then
the grouped convolution can be computed in parallel without any inter-processor communication.

Then each processor computes the global average pool of that result, yielding
a vector with one value for each channel. That vector multiplies the
corresponding block of the squeeze-matrix,
producing a partial sum of the squeeze-vector. The vectors produced by all the thread-blocks are added
in a buffer in global memory to produce the squeeze-vector. This addition in global memory
is a synchronization point for all the thread-blocks.

Then each processor loads the squeeze-vector from the global memory buffer,
applies the ReLU activation function, and multiples it by the block of the
excitement matrix corresponding to its channels. A sigmoid
activation is applied to the result of that transformation, producing gating values
between 0 and 1. These gating values multiply the corresponding channels of the
convolution layer output, producing the output of the Squeeze \& Excitation
layer (SE).

The SE output multiplies the corresponding block of weights from the second point-wise
convolution. The result is added on to the MBConv block's input tensor in global
memory, thereby producing the layer output with residual shortcut. The inter-block
synchronization in the SE layer also ensures that all thread-blocks have read
the input tensor before any of them update it with the MBConv output.

The block-fusion kernel computes the full height and width of the activation
tensors in each processor, effectively limiting the maximum resolution of the input images.
If the height and width are too large, then the blocks of the
activation tensors will not fit in local memory. We could extend the basic
block-fusion algorithm to partition the activation tensors in the height or
width dimensions in addition to the channels dimension and assign different blocks to different processors. This would
decrease the amount of local memory used by each processor, but it would also
require additional inter-processor communication to exchange the overlapping
``halos'' of the inputs to the grouped convolution layer. For example, each processor could
load an $h \times W \times C$ tile from the input tensor, where $h < H$, and compute the
first point-wise convolution for its channel group, yielding an  $h \times W \times r$ tensor, where $r < \alpha C$. Then each processor would exchange
its top and bottom rows with its neighbor, so that each now has a $(h + 2) \times W \times r$ tile
of hidden-layer activations. Then each processor computes the grouped convolution, producing an $h \times W \times r$ tile of activations.

The cost of this high-resolution block-fusion kernel is additional global memory
buffers, additional global memory data movement for halo exchange, and inter-processor synchronization. However
the extra data movement is much less than what is required to load the input
activation tensors and store the output activation tensors. We could increase
the number of hidden-layer channels computed by each processor and decrease the tile
size to keep the local memory allocation constant. This adjustment would increase the arithmetic intensity of the kernel
with respect to the global memory system.

The speed of inter-processor communication and synchronization can be relatively fast if processors are organized to support efficient communication with their neighbors. The distributed shared memory in the NVIDIA H100 GPU is one such example~\cite{nvidia2022h100}.

\FloatBarrier

\subsubsection{Waterline Analysis of MBConv}

\begin{figure}[h]
  \includegraphics[width=\textwidth]{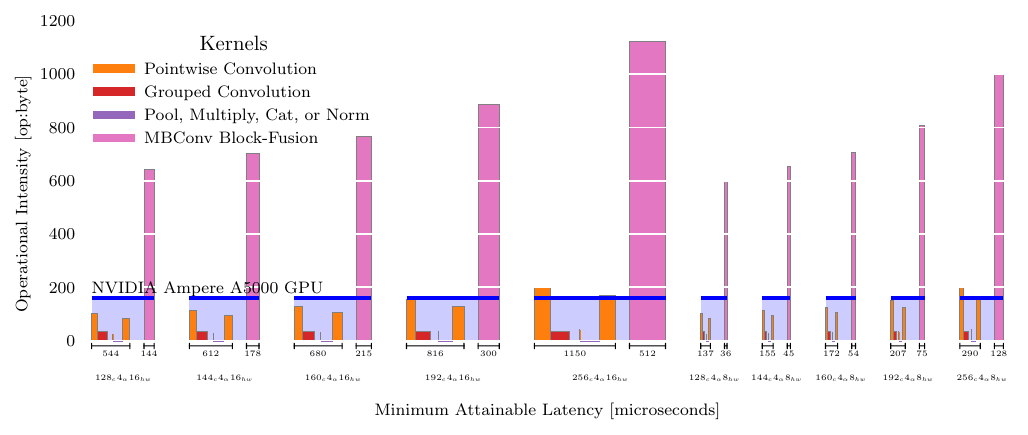}
  \caption{\textbf{Waterline plot} for the MBConv block with single-layer and block-fusion kernels. Single-layer kernels are memory-bound for grouped convolution, Squeeze \& Excitation layers, and some point-wise convolutions. The block-fusion kernels are strongly compute bound.}
  \label{sf:mbconv-waterline}
\end{figure}

Figure \ref{sf:mbconv-waterline} shows the waterline analysis of the MBConv single-layer and block-fusion kernels across a range of block hyperparameters. As mentioned, these kernels were designed for low-resolution feature maps found in the late stages of a convnet, so we use $16_H \times 16_W$ and $8_H \times 8_W$. Late vision blocks also have a relatively large number of channels, so we use $C=128$ and $C=256$. We also use bottleneck expansion ratio equal to four.

Point-wise convolution kernels are memory bound with 128 channels and are barely compute bound with 256 channels. The grouped convolution layers are also memory bound, as well as the SE layers.

The block-fusion kernels have very large operational intensity and are strongly compute bound. The minimum attainable latency of the block-fusion kernels is about one half to one third that of layer-by-layer execution.

\FloatBarrier

\subsection{CUDA Kernels}
\label{s:cuda-kernels}

\lstset{
  basicstyle=\footnotesize\ttfamily,
  breaklines,
  breakatwhitespace,
}

We implemented the ConvFirst and MBConv block-fusion kernels in CUDA~\cite{nvidiacuda}. We used code generation to
create helper classes to simplify tensor indexing. These classes compute the
offset for a given index using named-dimensions. For example, the tensor
spec
\begin{lstlisting}
    OutputIdx=dict(k8=8, n=128, p=64, q=64, k2=4)
\end{lstlisting}
generates a CUDA class that can be used like this:
\begin{lstlisting}
    st_outputs[OutputIdx().p(thread_p).q(thread_q).k8(k8)] =  val;
\end{lstlisting}

The indexing methods use \lstinline[language=C]!constexpr! so that the compiler
can simplify the indexing arithmetic at compile time. We found that the use of
dimension names when accessing high-dimensional tensors greatly improved the
clarity of the source code without sacrificing computational efficiency.

We also used code generation to implement classes that define static kernel
parameters, such as tile sizes. This enabled us to enumerate many kernel
configurations in Python without adding to the complexity of the CUDA source
code.

To launch our block-fusion kernels from Python, we implemented a \lstinline[language=C]!Kernel!
base class that uses the CuPy~\cite{cupy_learningsys2017} package to compile and launch kernels. We
implemented an auto-tune method that enumerates all valid kernel configurations
for the given block parameters, runs each with random weights and inputs, and selects the fastest.

To integrate our kernels with PyTorch~\cite{ansel2024pytorch}, we implemented a \lstinline[language=C]!TorchBlock! base
class that derives from \lstinline[language=C]!torch.nn.Module!. Each \lstinline[language=C]!TorchBlock!
derived class has a corresponding \lstinline[language=C]!Kernel! class. The first call to the
\lstinline[language=C]!TorchBlock.forward! method calls \lstinline[language=C]!Kernel.auto_tune! to select
the fastest kernel configuration for the block parameters. The kernel is then stored in
the block object and launched using CuPy. We use CuPy to capture a CUDA graph and
use the graph in the benchmarks to reduce kernel launch overhead.

This approach gives us a simple way to implement a custom inference backend that
runs in PyTorch. In principle, we could run an entire network this way. In this
paper, we simply benchmark a stage composed of multiple ConvFirst or MBConv blocks.

\subsubsection{ConvFirst CUDA Kernel}

The ConvFirst block-fusion kernel uses thread-blocks with eight warps (i.e.,
hardware-threads). The kernel computes a $p \times q \times K$ output tile by
first loading the corresponding $(p + 2) \times (q + 2) \times C$ input tile
into shared local memory (smem), where $C$ and $K$ are the number of input and
output channels, and $p$ and $q$ are the height and width of the output tile. The kernel
uses \lstinline[language=C]!float16! arithmetic with NVIDIA tensor core acceleration.

The warps compute the groups of the convolution layer in parallel, using the NVIDIA
\lstinline[language=C]!mma.m16n8k8.f16! matrix multiply instruction, which fits the group-width
of eight, and store the $p \times q \times C$ output to shared local memory. If $C == K$ and \lstinline[language=C]!stride == 1!, the block has a residual shortcut connection,
and each warp simultaneously loads a tile of $\frac{p \times q}{8} \times C$
input values from smem and uses them to initialize the registers that accumulate
 the second linear transformation of the FFN layer. Thus the
block input is added to the output during the first iteration of the FFN layer's multiply-accumulate operations.

More significantly, the residual shortcut does not cause an extra load from the
global memory system, as it would in layer-by-layer execution. Instead, the block-fusion kernel loads
the shortcut activations from smem, where they were previously stored for the grouped convolution.

Now that the grouped-convolution layer is computed and the FFN accumulators are
initialized, we are ready to compute the FFN layer. Each warp loads a $\frac{p \times q}{8} \times C$ tile of the grouped-convolution outputs into registers. Then each warp uses Equation~\eqref{e:ffn-eff} to compute a small number of hidden channels $r$, convert the \lstinline[language=C]!float32! output to \lstinline[language=C]!float16!, add bias and apply the ReLU activation function, and reformat the values as input fragments for the \lstinline[language=C]!mma.m16n8k8.f16! instruction. Then the fragments are multiplied by the corresponding weights of the second linear transform in the FFN layer with the result accumulated onto the output tile. Note that the hidden-layer activations never leave the register file.

The kernel loops over $r$ hidden channels at a time, asynchronously loading the
corresponding columns and rows of the FFN layer weights matrices at each
iteration. After the last iteration, the kernel adds the FFN layer's output bias
to the accumulators, converts them from \lstinline[language=C]!float32! to \lstinline[language=C]!float16!,
and stores the result to the output tile in global memory.

Ideally $\frac{p \times q}{8}$ is a multiple of 16 and $r$ is a multiple of 8 so
that the hidden activations tile with shape $\frac{p \times q}{8} \times r$ fits
the NVIDIA \lstinline[language=C]!mma.m16n8k8.f16! or \lstinline[language=C]!mma.m16n8k16.f16! matrix multiply instruction. ${p \times q}$ should be large enough to hide the latency of loading the weights from L2 cache, and $\frac{p \times q}{8}$ should be large enough to hide the latency of loading the weights from smem.

Of course $\frac{p \times q}{8} \times C$ input values of type
\lstinline[language=C]!float16! and $\frac{p \times q}{8} \times K$ output
accumulators of type \lstinline[language=C]!float32! must fit in each warp's
registers, in addition to a smaller number of registers for the weights and accumulators of the hidden-layer
activations. With a budget of 128 registers per warp,
our CUDA kernel performs well with up to 96 channels. This provides sufficient
width for the early stages of all but the largest convnets.

The block-fusion and scaling kernel shown in Figure
\ref{sf:convfirst-block-fusion-scaling} would support an arbitrary number of
channels, but we did not require it for the networks in this paper.

\subsubsection{MBConv CUDA Kernel}
\label{s:mbconv-cuda-kernel}

We wrote a CUDA kernel that implements the MBConv block. It uses the block-fusion algorithm defined by the tensor machine shown in Figure~\ref{f:mbconv-block-fusion-tensor-machine}. The kernels uses \lstinline[language=C]!float16! arithmetic with NVIDIA tensor core acceleration.

The CUDA kernel uses multiple thread-blocks, assigning a group of $64$
hidden-layer channels to each. Each thread-block corresponds to a single processor in Figure \ref{f:mbconv-block-fusion-tensor-machine}.

We used atomic adds to implement the additions on buffers in global memory. On NVIDIA GPUs, the global memory is the L2 cache, and atomic counters are used for
inter-thread-block synchronization.

Each thread-block computes the full height and width of the activation tensor.
Therefore, the kernel computes the convolution layer and global average
pooling without any inter-block communication. The maximum resolution that can be supported is limited by the requirement that the $H \times W \times 64_C$ tile of hidden-layer
activations must fit in shared local memory. We tested our kernels with $16_H \times 16_W$ and $4_N \times 8_H \times 8_W$ tensors, where $N$ is the batch
dimension, so that in both cases the kernel uses 32 KB of shared local memory
for activations. This fits easily in the 48 KB of smem available to each
thread-block, leaving room for a workspace used for loading weights tensors.

We could support high resolution by assigning spatial partitions of the activation tensor
to different thread-blocks. Additional inter-block communication would be needed to compute
the convolution and SE layers. It is unclear how much slowdown this extra communication
would cause on the Ampere GPU.

Inter-block communication is considerably faster on the next generation Hopper
GPU using distributed shared memory. Thread-blocks are organized into clusters
that can access each other's shared memory, enabling inter-block communication
that is about $7 \times$ faster than global memory~\cite{nvidia2022h100}.
Therefore, the Hopper GPU should make the MBConv kernel more efficient and
scalable.

We additionally implemented a \textit{micro-batching} optimization for the MBConv
kernel. Given a stage of $D$ consecutive MBConv blocks with feature-map size $H \times W$,
we compute the micro-batch size $n < N$ that is required to
fully-occupy all 64 streaming multi-processors on the A5000 GPU. Our kernel
computes the first $2 n$ images for the first MBConv block in the stage, writing
the output to global memory. Then the kernel advances to the next block in the stage, using the
$2 n$ outputs from the previous block as input. The kernel iterates over $d \leq D$ MBConv blocks
before looping back to the first block and processing the next micro-batch of $2 n$ images.

The idea is that the $2 n H W C$ bottleneck activations of the micro-batch and the approximately $d (2 \alpha C^2 + 72 \alpha C)$
weights for the point-wise and grouped-conv2d layers (of the $D$ MBConv blocks with bottleneck expansion ratio $\alpha$) fit in the L2
cache, so that intermediate activations are not stored to DRAM. Only the input and output
to the stack of $d$ blocks are transferred through DRAM. This works because the product
$H W C$ is relatively small for the late stages of the convnet where we use MBConv
blocks. Micro-batching increases the operational intensity of the MBConv kernel
until DRAM memory bandwidth has little impact on kernel performance.

The micro-batch optimization benefits from the fact that the block-fusion kernel
does not write hidden-layer activations to global memory, so the number of
activations that need to be cached is reduced by a factor of $\alpha + 1$.

\subsubsection{Kernel Benchmarks}

We benchmarked our kernels in PyTorch using the kernel launch method described
in Section~\ref{s:cuda-kernels}. We used CUDA events to record the start and end
time. We skipped 20 iterations of warmup to hide the auto-tune and ensure
stability of the GPU clock speed. We timed 100 iterations of the stage execution to get an accurate estimate of
the average latency. We benchmarked a stage of eight layers to ensure that
weights were not cached between iterations.

We benchmarked PyTorch Inductor by compiling the block module with \lstinline[language=C]!torch.compile()! and running with \lstinline[language=C]!torch.no_grad()!
and \lstinline[language=C]!torch.cuda.amp.autocast(dtype=torch.float16)! contexts. We again used warmup iterations to hide any recompilation
that might occur on the first iterations.

Our ConvFirst block-fusion kernel runs between $4.5 \times$ and $14.2 \times$
faster than PyTorch Inductor across a range of block configurations. The speedup
is greatest when the block is narrowest, confirming the intuition that the
single-layer kernels used by Inductor are memory-bound and become more efficient
as the operational intensity of the layers increases.

The computational efficiency of the block-fusion kernel can be as low as $35\%$
when the block is very narrow and the bottleneck expansion ratio is small.
However, when the number of channels reaches $32$ and the expansion ratio equals
$6$, the kernel is compute bound, and the computational efficiency equals
$66.9\%$, a $13.2 \times$ speedup over Inductor. With $64$ channels, our kernel
reaches $76.2\%$ computational efficiency.

\begin{figure}[h]
  \includegraphics[scale=0.8]{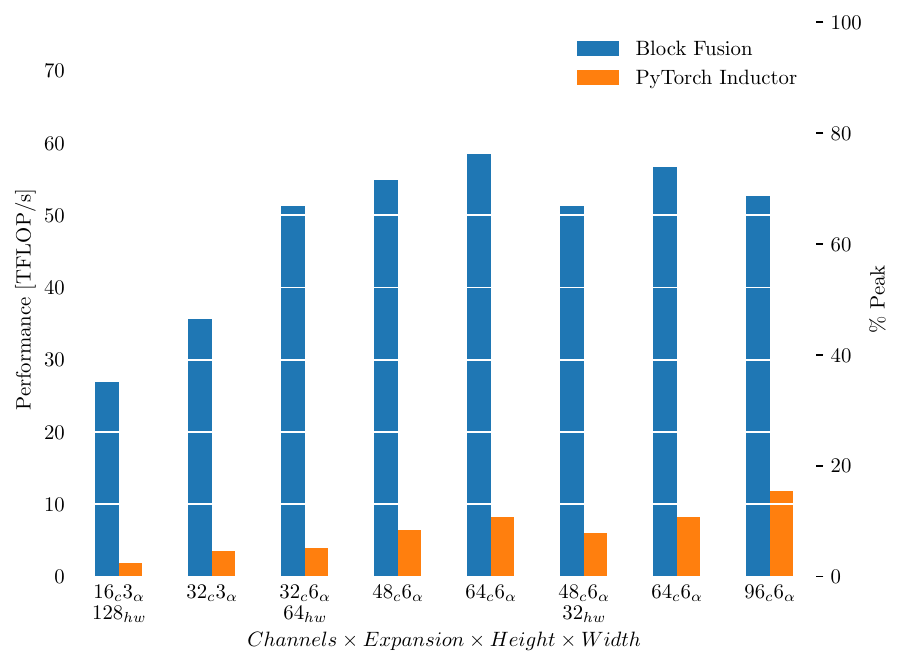}
  \caption{\textbf{ConvFirst block performance} in PyTorch, comparing the Inductor backend with our block-fusion kernels. The number of channels $C$, bottleneck expansion ratio $\alpha$, and feature map height and width $h w$ were selected from our ConvFirst network, and batch size equals $128$. Inductor's layer-by-layer operation schedule is memory bound, performing poorly when the number of channels is small and improving gradually as the number increases. Our block-fusion algorithm has greater operational intensity and therefore runs $14\times$ faster than Inductor when the number of channels is small and reaches $76\%$ of peak arithmetic throughput with 64 channels.}
  \label{f:convfirst_speed}
\end{figure}

\begin{table}[h]
  \scriptsize
  \caption{ConvFirst performance with PyTorch-Inductor versus our block-fusion kernels for different numbers of channels, bottleneck expansion ratio ($\alpha$), and activation tensor size (Height $\times$ Width).}

\begin{tabular}{ccrr|rrr|rrr|r}
\toprule
\multicolumn{4}{c}{} &  \multicolumn{3}{|c|}{Inductor} &  \multicolumn{3}{c|}{Block-Fusion} &   \\
Channels &  $\alpha$ &  Height & Width &  Time[ms] &  TFLOPS &  \% Peak &  Time[ms] &  TFLOPS &  \% Peak &  Speedup \\
\midrule
16 &  3 &   128 &      128 &               5.88 &             1.9 &           2.5 &           0.42 &        26.9 &      35.1 &    14.2 \\
32 &  3 &   128 &      128 &               9.89 &             3.6 &           4.7 &           0.99 &        35.7 &      46.6 &     9.9 \\
32 &  6 &    64 &      64 &               3.94 &             3.9 &           5.1 &           0.30 &        51.3 &      66.9 &    13.2 \\
48 &  6 &    64 &      64 &               5.06 &             6.4 &           8.4 &           0.59 &        54.9 &      71.6 &     8.6 \\
64 &  6 &    64 &      64 &               6.75 &             8.3 &          10.9 &           0.96 &        58.4 &      76.2 &     7.0 \\
48 &  6 &    32 &      32 &               1.33 &             6.1 &           8.0 &           0.16 &        51.3 &      67.0 &     8.4 \\
64 &  6 &    32 &      32 &               1.70 &             8.3 &          10.8 &           0.25 &        56.7 &      74.0 &     6.8 \\
96 &  6 &    32 &      32 &               2.60 &            11.8 &          15.4 &           0.58 &        52.6 &      68.7 &     4.5 \\
\bottomrule
\end{tabular}
  \label{t:conv-first-speed}
\end{table}

Our MBConv block-fusion kernel runs between $3.0 \times$ and $5.4 \times$ as
fast as PyTorch Inductor. The computational efficiency reaches $46.2\%$ with
$128$ channels and $51.4\%$ with 256 channels. Varying the height and width of the
activation tensor between $16 \times 16$ and $8 \times 8$ causes little
difference in performance. Performance drops when the number of channels is not
a power of two. This is caused by our kernel using a number of thread-blocks that does not evenly divide the GPU's $64$ available SMs.

The MBConv kernel uses very little DRAM bandwidth. Performance is limited by
memory latency and the speed of inter-block synchronization. The kernel would be
more efficient on a computer that affords fast inter-processor communication,
such as the NVIDIA Hopper architecture with distributed shared memory.

\begin{figure}[h]
  \includegraphics[scale=0.8]{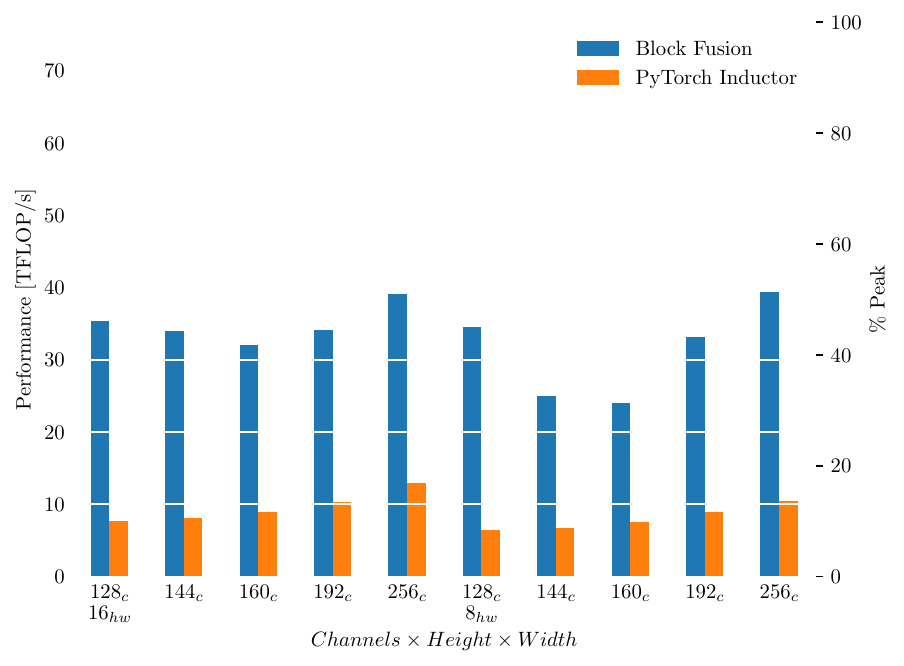}
  \caption{MBConv block performance with PyTorch-Inductor versus our block-fusion kernels. We varied the number of channels, height and width over the range of parameters used in our ConvFirst network.
    Bottleneck expansion ratio equals $4$ and batch size equals $128$ for all experiments.
    Inductor's layer-by-layer execution is severely memory bound for some layers of the MBConv block,
    while our fused-layer algorithm has large operational intensity. Inter-block synchronization proves costly for our kernels, limiting performance to $51\%$ of peak arithmetic throughput.
    Our kernels still run $3\times$ to $5.4\times$ faster than Inductor.
  }
  \label{f:mbconv_speed}
\end{figure}

\begin{table}[h]
  \scriptsize
  \caption{MBConv performance with PyTorch-Inductor versus our block-fusion kernels.}
  \begin{tabular}{ccrrr|rrr|rrr|r}
    \toprule
    \multicolumn{5}{c}{} &  \multicolumn{3}{|c|}{Inductor} &  \multicolumn{3}{c|}{Block-Fusion} &   \\
    Channels & $\alpha$ & Height & Width    & SE Ratio   &     Time[ms] &          TFLOPS &       \% Peak &      Time[ms] &      TFLOPS &  \% Peak &  Speedup \\
    \midrule
    128 &  4 &    16 &      16 &      0.25 &              11.46 &             7.7 &          10.0 &           2.48 &        35.4 &      46.2 &     4.6 \\
    144 &  4 &    16 &      16 &      0.25 &              13.38 &             8.1 &          10.6 &           3.19 &        34.0 &      44.4 &     4.2 \\
    160 &  4 &    16 &      16 &      0.25 &              14.81 &             8.9 &          11.6 &           4.09 &        32.1 &      41.9 &     3.6 \\
    192 &  4 &    16 &      16 &      0.25 &              17.78 &            10.3 &          13.5 &           5.39 &        34.1 &      44.4 &     3.3 \\
    256 &  4 &    16 &      16 &      0.25 &              24.30 &            12.9 &          16.8 &           8.01 &        39.1 &      51.0 &     3.0 \\
    128 &  4 &     8 &       8 &      0.25 &               3.43 &             6.4 &           8.4 &           0.64 &        34.5 &      45.0 &     5.4 \\
    144 &  4 &     8 &       8 &      0.25 &               4.03 &             6.7 &           8.8 &           1.09 &        25.0 &      32.5 &     3.7 \\
    160 &  4 &     8 &       8 &      0.25 &               4.36 &             7.5 &           9.8 &           1.37 &        24.0 &      31.3 &     3.2 \\
    192 &  4 &     8 &       8 &      0.25 &               5.18 &             8.9 &          11.6 &           1.39 &        33.1 &      43.1 &     3.7 \\
    256 &  4 &     8 &       8 &      0.25 &               7.56 &            10.4 &          13.5 &           1.99 &        39.4 &      51.4 &     3.8 \\
    \bottomrule
  \end{tabular}
  \label{t:mbconv-speed}
\end{table}
\FloatBarrier

\subsection{Depth-First Execution}

Many papers have pursued the idea of depth-first execution of
convnets~\cite[\dots]{alwani2016fused, goetschalckx2019breaking}. The idea is to compute only a small tile of activations in the first
layer, then immediately compute all the outputs in the next layer that can be
produced from it. By chaining several layers together in this fashion, one can
compute several layers at a time while keeping the activation tiles small enough to fit in
on-chip memory.

Global operators like Squeeze-and-Excitation prevent depth-first
execution. Because every output of these layers depends on every input, there is
no depth-first partition of the computational graph. Therefore, depth-first execution cannot
be applied to the MBConv+SE blocks used in this paper.

But depth-first execution complements the ConvFirst block-fusion kernels. One
would consider a ConvFirst block as a specialized $3 \times 3$ convolutional
layer with $N H W C$ input and output tensors and large arithmetic intensity
with respect to its weights. The dependency analysis for partitioning the activation
tensors into smaller tiles is the same as for a regular $3 \times 3$ convolution. Therefore, we
can enjoy the small memory footprint of a depth-first execution across a stack of ConvFirst blocks.

Without block-fusion kernels, the ConvFirst blocks appear as a grouped
convolution followed by two point-wise convolutions. Each of these separately
has much smaller arithmetic intensity than the block-fusion kernel, so
depth-first execution of un-fused layers would require more data movement in the
on-chip global memory system. Also it would require additional
global memory for storing tiles of the hidden-layer activations. Therefore, we can minimize
data movement at all layers of the memory hierarchy by using depth-first execution in conjunction with block-fusion kernels.

Additionally, depth-first execution requires a tile schedule that spans multiple
layers with some synchronization between tiles. This might be difficult to map
to the CUDA programming model and the large parallelism of GPUs. On the other hand,
it could be very efficient on small accelerators where the memory efficiency would
prove most beneficial.

\section{ConvFirstNet: Striving for Computational Efficiency and Model Efficiency}\label{s:section}
\label{s:ConFirstNet}

With the goal of co-optimizing model efficiency and computational
efficiency, we used the ConvFirst and MBConv blocks to design a convnet model called
ConvFirstNet.

\subsection{The Design of ConvFirstNet}

 We used ConvFirst blocks for the early stages (1-3) and MBConv
blocks for the late stages (4-5). This division between light-weight early
vision stages and high-capacity late vision stages is inspired by
EfficientNetV2~\cite{tan2021efficientnetv2}. However, our ConvFirst block uses many fewer operations and
parameters than the FusedMBConv block used by EfficientNetV2.

Similar to NFNet~\cite{brock2021high}, ConvFirstNet increases the number of channels between the
third and fourth stages by a large factor. Unlike other models, ConvFirstNet uses the same
number of blocks in the fourth and fifth stages.

We used group-width equal to eight for all grouped convolutions because it is
the smallest width that is divisible by the NVIDIA
\lstinline[language=C]!mma.m16n8k8.f16! matrix-multiply instruction.
In CIFAR experiments, we found group-width equal to four yielded the best model
efficiency. Eight and one (depth-wise) both used slightly more
operations to achieve the same accuracy. We would prefer group-width four on
hardware that computed it efficiently.

We used an expansion ratio equal to three in the bottleneck of the first ConvFirst block
to reduce the amount of computation done by this stage. This might have
been a mistake, because a larger expansion ratio would increase the operational
intensity and yield better computational efficiency. Another interesting idea is to replace the stem and first stage
with a \textit{patchify} layer with stride equal to four, as ConvNeXt
does.

Stages two and three use ConvFirst with expansion ratio equal to six. This
is inspired by the large expansion ratios used by EfficientNet's bottleneck blocks.

Stages four and five use MBConv with expansion ratio equal to four. We
chose this smaller ratio to decrease the number of operations performed
by each block, because setting group-width equal to eight increased the number of operations
relative to the typical MBConv block with depth-wise convolution.

The MBConv block has a Squeeze \& Excitation layer with squeeze-ratio equal to
0.25, same as EfficientNet.

The complete model definitions are listed in Table \ref{t:convfirstnet-small}.
These architecture hyperparameters worked best across a small range of tested
models. The choices are undoubtedly suboptimal and could be improved by a
thorough search of the design space. In particular, the ConvFirst-Small model is
likely too wide and shallow. Simply increasing the depth of ConvFirstNet-Small
might produce an efficient ConvFirstNet-Medium.

All models use the same $256 \times 256$ image resolution. The power-of-two size
ensures that the activation tensors at every stage are evenly divisible by
the matrix fragments used by NVIDIA tensor cores.

\begin{table}[h]
  \scriptsize
  \caption{\textbf{ConvFirstNet} uses narrow, computationally efficient ConvFirst blocks in the early stages of the network and wide, high capacity MBConv blocks in the late stages. The division between early and late stage vision was inspired by EfficientNetV2. All Conv2d layers use $3 \times 3$ kernels. In the ConvFirst and MBConv blocks, the Conv2d layers also use group-width equal to 8, which matches the dimensions of the NVIDIA tensor core \lstinline[language=C]!mma.m16n8k8! instruction. The point-wise convolution layers do not use groups. ConvFirst6 is a ConvFirst block with 6 times as many hidden channels as input channels, and MBConv4 is an MBConv block with $4 \times$ expansion. MBConv blocks have Squeeze \& Excitation layers with squeeze-ratio equal to 0.25. The Conv layer stem and ConvFirst blocks use ReLU activations and MBConv uses SiLU.}
  \begin{tabular}{c|l|c|r|r}
    \toprule
    \multicolumn{5}{c}{ConvFirstNet-Pico}                         \\
    \midrule
    Stage & Block               & Stride & \#Channels & \#Blocks \\
    \midrule
    0     & Conv                   &   2    &    16     &    1    \\
    1     & ConvFirst3             &   1    &    16     &    1    \\
    2     & ConvFirst6             &   2    &    32     &    2    \\
    3     & ConvFirst6             &   2    &    48     &    3    \\
    4     & MBConv4, SE0.25        &   2    &   128     &   11    \\
    5     & MBConv4, SE0.25        &   2    &   128     &   11    \\
    Head  & Conv1x1, Pool, FC      &        &  1280     &   1     \\
    \toprule
    \multicolumn{5}{c}{ConvFirstNet-Nano}                         \\
    \midrule
    Stage & Block               & Stride & \#Channels & \#Blocks \\
    \midrule
    0     & Conv                   &   2    &    24     &    1    \\
    1     & ConvFirst3             &   1    &    24     &    1    \\
    2     & ConvFirst6             &   2    &    48     &    3    \\
    3     & ConvFirst6             &   2    &    64     &    4    \\
    4     & MBConv4, SE0.25        &   2    &   160     &   14    \\
    5     & MBConv4, SE0.25        &   2    &   160     &   14    \\
    Head  & Conv1x1, Pool, FC      &        &  1280     &   1     \\
    \toprule
    \multicolumn{5}{c}{ConvFirstNet-Tiny}                         \\
    \midrule
    Stage & Block               & Stride & \#Channels & \#Blocks \\
    \midrule
    0     & Conv                   &   2    &    24     &    1    \\
    1     & ConvFirst3             &   1    &    24     &    2    \\
    2     & ConvFirst6             &   2    &    48     &    5    \\
    3     & ConvFirst6             &   2    &    72     &    6    \\
    4     & MBConv4, SE0.25        &   2    &   192     &   18    \\
    5     & MBConv4, SE0.25        &   2    &   192     &   18    \\
    Head  & Conv1x1, Pool, FC      &        &  1280     &   1     \\
    \toprule
    \multicolumn{5}{c}{ConvFirstNet-Small}                         \\
    \midrule
      Stage & Block               & Stride & \#Channels & \#Blocks \\
      \midrule
      0     & Conv                   &   2    &    32     &    1    \\
      1     & ConvFirst3             &   1    &    32     &    2    \\
      2     & ConvFirst6             &   2    &    64     &    5    \\
      3     & ConvFirst6             &   2    &    96     &    6    \\
      4     & MBConv4, SE0.25        &   2    &   256     &   18    \\
      5     & MBConv4, SE0.25        &   2    &   256     &   18    \\
      Head  & Conv1x1, Pool, FC      &        &  1280     &   1     \\
      \bottomrule
    \end{tabular}
    \label{t:convfirstnet-small}
\end{table}

\subsection{Waterline Analysis of ConvFirstNet}

\begin{figure}[h]
  \includegraphics[width=\textwidth]{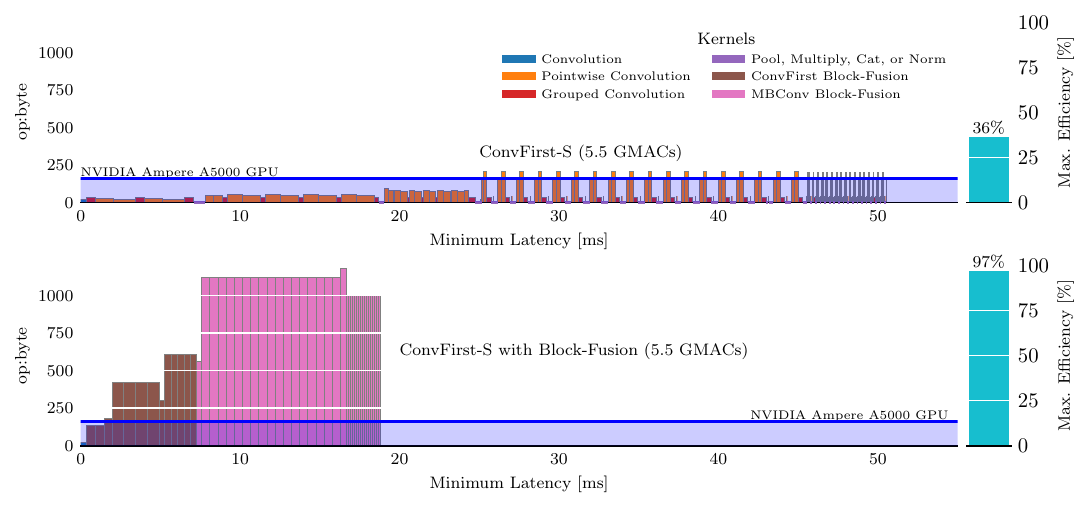}
  \caption{\textbf{Waterline analysis} of ConvFirst models. Layer-by-layer execution
    makes ConvFirstNet memory bound and severely limits its maximum computational
    efficiency. Block-fusion kernels make almost the entire network compute
    bound and near perfect computational efficiency is attainable.}
  \label{f:waterline-convfirstnet}
\end{figure}

Figure \ref{f:waterline-convfirstnet} compares the waterline analyses of
ConvFirstNet-Small using single-layer and block-fusion kernels. Similar to EfficientNet,
ConvFirst-Small is severely memory-bound with layer-by-layer execution. The ConvFirst blocks
in the early stages and the grouped convolutions and Squeeze \& Excitation
layers in the late stages all have very low operational intensity relative to
the A5000 op:byte ratio. The resulting maximum computational efficiency equals
$36\%$.

Block-fusion kernels change the picture dramatically. All but the first stage
have operational intensity high above the waterline. The
resulting maximum computational efficiency equals $97\%$.

Figure~\ref{f:waterline-convfirstnet} illustrates the peril of using an
off-the-shelf inference engine to measure the computational efficiency of a
model. Had we used layer-by-layer execution to evaluate the efficiency of ConvFirstNet,
we would have judged it to be inferior and discarded it. But ConvFirstNet can attain very high computational efficiency when
computed with block-fusion kernels.

\begin{figure}[h]
  \includegraphics[width=0.6\textwidth]{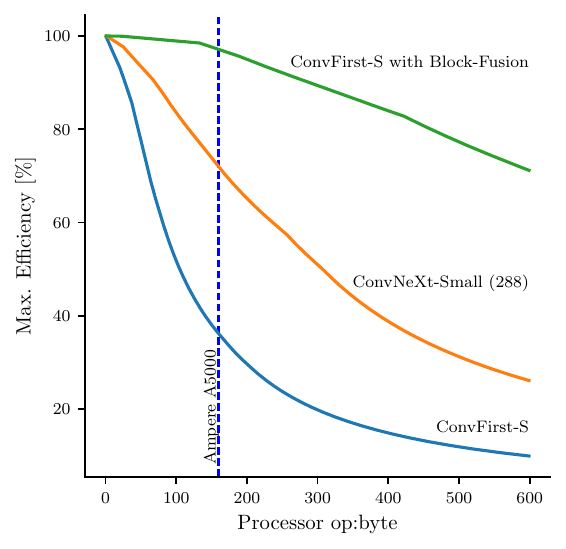}
  \caption{Maximum attainable computational efficiency of ConvFirstNet and ConvNeXt over a range of processor op:byte ratios. Block fusion kernels make the networks efficient even when the processor has significantly greater op:byte ratio than our baseline A5000 GPU. Layer-by-layer execution requires very low op:byte ratios for efficient operation.}
  \label{f:rising-water}
\end{figure}

Figure~\ref{f:rising-water} shows the effect that increasing op:byte ratio has
on maximum attainable efficiency. Block-fusion kernels enable ConvFirst
to perform well with relatively little DRAM bandwidth; 80\% of peak arithmetic
throughput is still attainable when the processor op:byte ratio is 500.
Layer-by-layer execution causes both ConvFirst and ConvNeXt to be slow unless the processor has a very low op:byte ratio.

\FloatBarrier

\subsection{ConvFirstNet Model Experiments}

We implemented the ConvFirst model in PyTorch~\cite{ansel2024pytorch} and trained it using the PyTorch Image Models framework~\cite{rw2019timm}. We trained for 600 epochs with \lstinline[language=C]!float16! precision using \lstinline[language=C]!native-amp! and an initial learning rate of $0.128$ with 3 epochs of warmup~\cite{goyal2017accurate}. We used image size equal to $224 \times 224$ for training and $256 \times 256$ for inference for all model sizes.  We trained with eight NVIDIA A6000 GPUs, each with batch size 256, for a total batch size of 2048. Momentum was equal to 0.9 and weight-decay was 1.0e-05. We used the RMSprop optimizer with ``TensorFlow style epsilon'', decay epochs equal to 2.4, and decay rate equal to 0.97. These optimizer settings were inspired by EfficientNet. ConvFirst-Pico used dropout equal to 0.2 and the rest used 0.3. All models used drop-path (i.e., stochastic depth~\cite{huang2016deep}) equal to 0.2. We used data augmentation with the auto-augment setting \lstinline[language=C]!rand-m9-mstd0.5!, which corresponds to RandAugment~\cite{cubuk2020randaugment} with magnitude 9 and noise of standard deviation equal to 0.5. We also used random-erase~\cite{zhong2020random} in ``pixel'' mode with probability equal to 0.2. We used an exponential moving average of the weights.

We benchmarked the baseline models using the implementations included with PyTorch Image Models and the \lstinline[language=C]!benchmark.py! script. We used precision \lstinline[language=C]!float16! and \lstinline[language=C]!--channels-last!, \lstinline[language=C]!--torchcompile=inductor!, \lstinline[language=C]!--bench=inference!, and set the batch size equal to 128. We set the GPU and memory clocks to the same values used in the block-fusion benchmarks in the previous section.

We did not implement block-fusion kernels for all the blocks of ConvFirstNet.
Specifically, ConvFirst and MBConv with stride 2, ConvFirst with 24 channels,
and the stem convolution were not implemented. However the unimplemented layers
account for a small fraction of the total network computation, so we can reasonably
estimate the performance of the entire network using the measured latency for
the implemented blocks and estimates for the rest. See Appendix
\ref{a:appendix1} for details.

\subsection{Results}

The speed and accuracy results for ConvFirstNet with block-fusion kernels and baseline models with PyTorch Inductor are listed in Table \ref{t:models}.
ConvFirstNet with block-fusion has greater model efficiency than EfficientNet and greater computational efficiency than ConvNeXt with PyTorch Inductor.

\begin{figure}[h]
  \includegraphics[width=\textwidth]{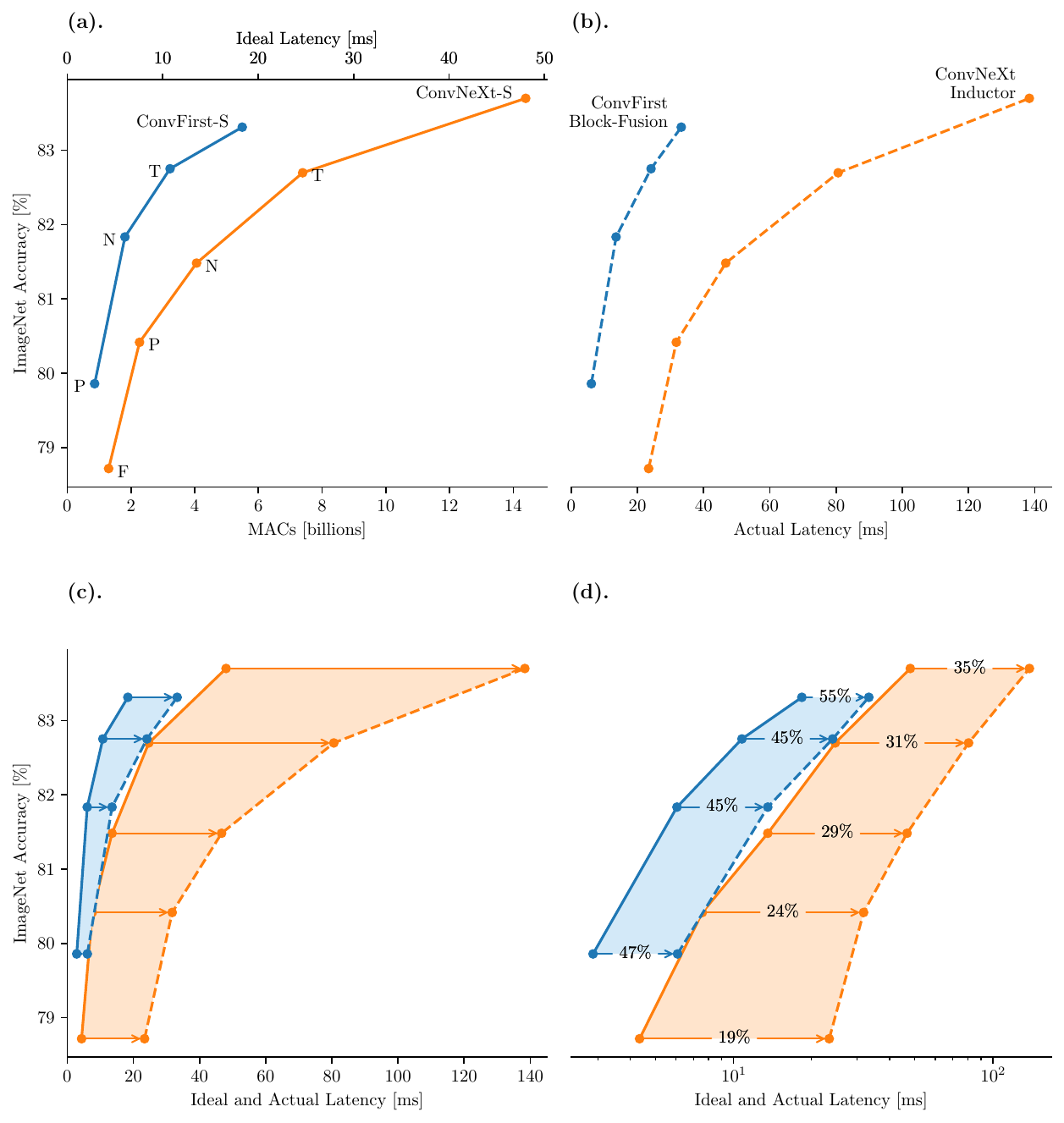}
  \caption{ConvFirst with block-fusion versus ConvNeXt with PyTorch Inductor.
    \textbf{(a).} Our ConvFirst model has significantly better model efficiency than ConvNeXt, achieving greater accuracy with fewer operations.
    \textbf{(b).} ConvFirst extends its lead when actual latency is measured with our block-fusion kernels.
    \textbf{(c).} The actual latency of ConvFirst is lower than the ideal latency of ConvNeXt.
    \textbf{(d).} The narrower efficiency gap of ConvFirst results from greater computational efficiency that ranges from 47\% -- 55\%.
  }
  \label{f:convfirst-convnext-gap}
\end{figure}

Figure~\ref{f:convfirst-convnext-gap} shows efficiency gap plots comparing ConvFirstNet and ConvNeXt. ConvFirstNet has a significant model
efficiency advantage over ConvNeXt, and the block-fusion kernels make
ConvFirstNet more computationally efficient also. The combination of efficient
model and operationally intense algorithms makes ConvFirstNet's actual latency
lower than ConvNeXt's ideal latency.

\begin{table}[h]
  \scriptsize
  \caption{\textbf{ConvFirstNet Speed and Accuracy.} Our ConvFirst model with block-fusion has greater model efficiency than EfficientNet and greater computational efficiency than ConvNeXt. The combined advantages of an efficient model with memory-efficient algorithms gives ConvFirst frame rates (FPS) approximately $4\times$ faster than the baseline at equal accuracy. We measured ImageNet-1K classification accuracy and latency using an NVIDIA Ampere A5000 GPU with $76.7$ TFLOP/s peak arithmetic throughput and batch size $128$. All models used PyTorch. ConvFirst used our block-fusion kernels and the rest used PyTorch Inductor.}
\begin{tabular}{lrrrrrr}
\toprule
           Model &  Params [M] &  MACs [B] &  Latency [ms] &     FPS &  \% Peak &  Accuracy [\%] \\
\midrule
ConvNeXt Femto (288) &         5.2 &       1.3 &          23.4 &  5,474.8 &    18.6 &          78.7 \\
 EfficientNet B1 &         7.8 &       0.8 &          49.9 &  2,566.8 &     5.2 &          79.1 \\
  \textbf{ConvFirst Pico} &         5.9 &       0.9 &           6.1 & \textbf{21,059.6} &    47.2 &          \textbf{79.9} \\
  \midrule
 EfficientNet B2 &         9.1 &       1.1 &          66.1 &  1,935.9 &     5.7 &          80.1 \\
 ConvNeXt Pico (288) &         9.1 &       2.3 &          31.7 &  4,033.3 &    23.9 &          80.4 \\
 ConvNeXt Nano (288) &        15.6 &       4.1 &          46.7 &  2,742.5 &    29.0 &          81.5 \\
 EfficientNet B3 &        12.2 &       2.0 &         107.5 &  1,190.4 &     6.2 &          81.6 \\
 \textbf{ConvFirst Nano} &        10.2 &       1.8 &          13.5 &  \textbf{9,447.2} &    44.6 &          \textbf{81.8} \\
 \midrule
   ConvNeXt Tiny &        28.6 &       7.4 &          80.6 &  1,587.8 &    30.6 &          82.7 \\
   \textbf{ConvFirst Tiny} &        17.2 &       3.2 &          24.1 &  \textbf{5,311.0} &    44.7 &          \textbf{82.8} \\
   \midrule
 EfficientNet B4 &        19.3 &       4.5 &         203.7 &   628.3 &     7.4 &          82.9 \\
 \textbf{ConvFirst Small} &        28.6 &       5.5 &          33.2 &  \textbf{3,851.0} &    55.2 &          \textbf{83.3} \\
 \midrule
 EfficientNet B5 &        30.4 &       9.6 &         385.1 &   332.4 &     8.3 &          83.6 \\
  ConvNeXt Small (288) &        50.2 &      14.4 &         138.4 &   925.1 &    34.7 &          83.7 \\
EfficientNetv2 S &        21.5 &       8.4 &         146.0 &   876.8 &    19.3 &          83.9 \\
\bottomrule
\end{tabular}
\label{t:models}
\end{table}

\FloatBarrier

\section{Conclusion}\label{s:ccl}

We presented a new convnet called ConvFirst that has greater model efficiency
than EfficientNet and greater computational efficiency than ConvNeXt, when implemented
with our block-fusion kernels.

We observed that modern convnets use blocks composed of
degenerate \lstinline[language=C]!conv2d! layers that have relatively low
operational intensity. Thus the traditional layer-by-layer execution of convnets
is memory bound with limited computational efficiency.

We proposed to make the convnet block the target of optimization and designed
block-fusion kernels that run significantly faster than single-layer kernels.

We also developed analytical tools to examine model efficiency. We derived the
simple formulas that show how model efficiency and computational efficiency
combine to produce latency. We created the efficiency gap plot to show model
efficiency and latency on the same axes, despite the prevailing view that these concepts are irreconcilable. We created waterline analysis to
study the maximum attainable efficiency of a sequence of parallel kernels. We
created tensor machines to illustrate the mechanics of block-fusion kernels.

We have shown that large improvements in convnet efficiency can be made by
identifying the separate model and computational sources of efficiency and
co-optimizing both. We have also shown that the actual performance of efficient models can
be severely memory bound when run with layer-by-layer execution. Therefore, it is no longer
acceptable for deep learning researchers to estimate the computational efficiency
of a model without also studying the efficiency of the algorithms that compute it.

We believe that deep learning must be studied not just as a mathematical
modeling problem, but also as a computer science problem. A model's strength is
determined not only by the accuracy that it achieves, but also by the
computational efficiency that it affords. By co-optimizing models and
algorithms, we can develop a new wave of convnets that deliver significantly greater performance.

\section*{Acknowledgment}
The author would like to thank Hyunggi Cho, CEO \& Co-Founder of Phantom AI. This project would not have been possible without his unwavering support.

\printbibliography

\appendix
\section{ConvFirstNet Speed Projections}\label{a:appendix1}

We did not implement all the ConvFirst and MBConv kernels that are required
to run ConvFirstNet. The stride-2 kernels, ConvFirst with channels
equal to 24, and the stem kernels are not implemented. Therefore, we estimate the
network inference time by using the benchmark results of the implemented
blocks and efficiency estimates of the unimplemented blocks. We estimate
that the stride-2 blocks will run at 80\% of the efficiency of the stride-1
blocks, that the ConvFirst block with 24 channels and expansion factor 3 will
run at 40\% computational efficiency, and the stem convolution will run at 75\%.

The stem has 3 input channels and 16, 24, or 32 output channels. It would be
fused with the kernel of the first ConvFirst block, significantly reducing the
size of the block's input tensor, likely causing the block to run faster. We feel this
justifies the relatively high efficiency estimate for the stem.

The unimplemented kernels are a relatively small fraction of the total
network, so inaccurate estimates will not greatly affect the accuracy of the
network run time projection.

The following tables contain the details of the ConvFirst speed projections.

\begin{sidewaystable}[h]
  \scriptsize
  \caption{\textbf{ConvFirstNet-Pico} speed projection.}
\begin{tabularx}{\textwidth}{lrrrrrrrrrr|rrrrrr|rXXX}
\toprule
      &   &        &    &   &     &    &   &   &     &    & \multicolumn{6}{c|}{Layer OPs [millions]}  &  & & \\
Block     & N   & s &  H  & W & Ks  & gw & t & C &  R  &  K & Conv & Exp & Prj & Sqz Exc & Depth & Total & \% Peak & Est. \%Peak & Latency [ms] & TFLOP/s \\
\midrule
Stem      & 128 & 2 & 256 & 256 & 3 & & & 3 & & 16 & 14.16 &  &  &  & 1 & 14.16 &  & 75.0\% & 0.031 & 57.5 \\
ConvFirst & 128 & 1 & 128 & 128 & 3 & 8 & 3 & 16 & 48 & 16 & 37.75 & 25.17 & 25.17 & & 1 & 88.08 & 35.1\% &  & 0.419 & 26.9 \\
ConvFirst & 128 & 2 & 128 & 128 & 3 & 8 & 6 & 16 & 96 & 32 & 37.75 & 25.17 & 25.17 & & 1 & 88.08 &  & 53.5\% & 0.275 & 41.0 \\
ConvFirst & 128 & 1 & 64 & 64 & 3 & 8 & 6 & 32 & 192 & 32 & 18.87 & 50.33 & 50.33 & & 1 & 119.54 & 66.9\% &  & 0.298 & 51.3 \\
ConvFirst & 128 & 2 & 64 & 64 & 3 & 8 & 6 & 32 & 192 & 48 & 18.87 & 25.17 & 18.87 & & 1 & 62.91 &  & 53.6\% & 0.196 & 41.1 \\
ConvFirst & 128 & 1 & 32 & 32 & 3 & 8 & 6 & 48 & 288 & 48 & 7.08 & 28.31 & 28.31 & & 2 & 127.40 & 67.0\% &  & 0.317 & 51.4 \\
MBConv    & 128 & 2 & 32 & 32 & 3 & 8 & 4 & 48 & 192 & 128 & 28.31 & 18.87 & 12.58 & 0.01 & 1 & 59.78 &  & 37.0\% & 0.270 & 28.3 \\
MBConv    & 128 & 1 & 16 & 16 & 3 & 8 & 4 & 128 & 512 & 128 & 18.87 & 33.55 & 33.55 & 0.07 & 10 & 860.49 & 46.2\% &  & 3.108 & 35.4 \\
MBConv    & 128 & 2 & 16 & 16 & 3 & 8 & 4 & 128 & 512 & 128 & 18.87 & 33.55 & 8.39 & 0.07 & 1 & 60.88 &  & 36.0\% & 0.282 & 27.6 \\
MBConv    & 128 & 1 & 8 & 8 & 3 & 8 & 4 & 128 & 512 & 128 & 4.72 & 8.39 & 8.39 & 0.07 & 10 & 215.61 & 45.0\% &  & 0.800 & 34.5 \\
\midrule
Total    &      &   &   &   &   &   &   &     &     &     &      &      &      &      &    & 1,696.93 &  & 47.2 & 5.996 & 36.2 \\
\bottomrule
\end{tabularx}
\label{t:convfirstnet-pico-speed}
\end{sidewaystable}

\begin{sidewaystable}[h]
  \scriptsize
  \caption{\textbf{ConvFirstNet-Nano} speed projection.}
  \begin{tabularx}{\textwidth}{lrrrrrrrrrr|rrrrrr|rXXX}
    \toprule
    &   &        &    &   &     &    &   &   &     &    & \multicolumn{6}{c|}{Layer OPs [millions]}  &  & & \\
    Block     & N   & s &  H  & W & Ks  & gw & t & C &  R  &  K & Conv & Exp & Prj & Sqz Exc & Depth & Total & \% Peak & Est. \%Peak & Latency [ms] & TFLOP/s \\
    \midrule
Stem & 128 & 2 & 256 & 256 & 3 & & & 3 & & 24 & 21.23 &  &  &  & 1 & 21.23 & 0.0\% & 75.0\% & 0.047 & 57.5 \\
ConvFirst & 128 & 1 & 128 & 128 & 3 & 8 & 3 & 24 & 72 & 24 & 56.62 & 56.62 & 56.62 &  & 1 & 169.87 & 0.0\% & 40.0\% & 0.709 & 30.7 \\
ConvFirst & 128 & 2 & 128 & 128 & 3 & 8 & 6 & 24 & 144 & 48 & 56.62 & 56.62 & 56.62 &  & 1 & 169.87 & 0.0\% & 57.3\% & 0.495 & 43.9 \\
ConvFirst & 128 & 1 & 64 & 64 & 3 & 8 & 6 & 48 & 288 & 48 & 28.31 & 113.25 & 113.25 &  & 2 & 509.61 & 71.6\% & 0.0\% & 1.188 & 54.9 \\
ConvFirst & 128 & 2 & 64 & 64 & 3 & 8 & 6 & 48 & 288 & 64 & 28.31 & 56.62 & 37.75 &  & 1 & 122.68 & 0.0\% & 61.0\% & 0.336 & 46.8 \\
ConvFirst & 128 & 1 & 32 & 32 & 3 & 8 & 6 & 64 & 384 & 64 & 9.44 & 50.33 & 50.33 &  & 3 & 330.30 & 76.2\% & 0.0\% & 0.723 & 58.4 \\
MBConv    & 128 & 2 & 32 & 32 & 3 & 8 & 4 & 64 & 256 & 160 & 37.75 & 33.55 & 20.97 & 0.02 & 1 & 92.29 & 0.0\% & 33.5\% & 0.459 & 25.7 \\
MBConv    & 128 & 1 & 16 & 16 & 3 & 8 & 4 & 160 & 640 & 160 & 23.59 & 52.43 & 52.43 & 0.10 & 13 & 1,671.19 & 41.9\% & 0.0\% & 6.656 & 32.1 \\
MBConv    & 128 & 2 & 16 & 16 & 3 & 8 & 4 & 160 & 640 & 160 & 23.59 & 52.43 & 13.11 & 0.10 & 1 & 89.23 & 0.0\% & 25.0\% & 0.595 & 19.2 \\
MBConv    & 128 & 1 & 8 & 8 & 3 & 8 & 4 & 160 & 640 & 160 & 5.90 & 13.11 & 13.11 & 0.10 & 13 & 418.80 & 31.3\% & 0.0\% & 2.233 & 24.0 \\
\midrule
Total &  &  &  &  &  &  &  &  &  &  &  &  &  &  &  & 3,595.07 & 44.6\% &  & 13.441 & 34.2 \\
\bottomrule
  \end{tabularx}
  \label{t:convfirstnet-nano-speed}
\end{sidewaystable}

\begin{sidewaystable}[h]
  \scriptsize
  \caption{\textbf{ConvFirstNet-Tiny} speed projection.}
  \begin{tabularx}{\textwidth}{lrrrrrrrrrr|rrrrrr|rXXX}
    \toprule
    &   &        &    &   &     &    &   &   &     &    & \multicolumn{6}{c|}{Layer OPs [millions]}  &  & & \\
    Block     & N   & s &  H  & W & Ks  & gw & t & C &  R  &  K & Conv & Exp & Prj & Sqz Exc & Depth & Total & \% Peak & Est. \%Peak & Latency [ms] & TFLOP/s \\
    \midrule
Stem & 128 & 2 & 256 & 256 & 3 & 0 & 0 & 3 & 0 & 24 & 21.23 & 0.00 & 0.00 & 0.00 & 1 & 21.23 & 0.0\% & 75.0\% & 0.047 & 57.5 \\
ConvFirst & 128 & 1 & 128 & 128 & 3 & 8 & 3 & 24 & 72 & 24 & 56.62 & 56.62 & 56.62 & 0.00 & 2 & 339.74 & 0.0\% & 40.0\% & 1.417 & 30.7 \\
ConvFirst & 128 & 2 & 128 & 128 & 3 & 8 & 6 & 24 & 144 & 48 & 56.62 & 56.62 & 56.62 & 0.00 & 1 & 169.87 & 0.0\% & 57.3\% & 0.495 & 43.9 \\
ConvFirst & 128 & 1 & 64 & 64 & 3 & 8 & 6 & 48 & 288 & 48 & 28.31 & 113.25 & 113.25 & 0.00 & 4 & 1,019.22 & 71.6\% & 0.0\% & 2.376 & 54.9 \\
ConvFirst & 128 & 2 & 64 & 64 & 3 & 8 & 6 & 48 & 288 & 72 & 28.31 & 56.62 & 42.47 & 0.00 & 1 & 127.40 & 0.0\% & 56.0\% & 0.380 & 43.0 \\
ConvFirst & 128 & 1 & 32 & 32 & 3 & 8 & 6 & 72 & 432 & 72 & 10.62 & 63.70 & 63.70 & 0.00 & 5 & 690.09 & 0.0\% & 70.0\% & 1.645 & 53.7 \\
MBConv & 128 & 2 & 32 & 32 & 3 & 8 & 4 & 72 & 288 & 192 & 42.47 & 42.47 & 28.31 & 0.02 & 1 & 113.27 & 0.0\% & 33.5\% & 0.564 & 25.7 \\
MBConv & 128 & 1 & 16 & 16 & 3 & 8 & 4 & 192 & 768 & 192 & 28.31 & 75.50 & 75.50 & 0.15 & 17 & 3,050.72 & 41.9\% & 0.0\% & 12.151 & 32.1 \\
MBConv & 128 & 2 & 16 & 16 & 3 & 8 & 4 & 192 & 768 & 192 & 28.31 & 75.50 & 18.87 & 0.15 & 1 & 122.83 & 0.0\% & 25.0\% & 0.819 & 19.2 \\
MBConv & 128 & 1 & 8 & 8 & 3 & 8 & 4 & 192 & 768 & 192 & 7.08 & 18.87 & 18.87 & 0.15 & 17 & 764.56 & 31.3\% & 0.0\% & 4.076 & 24.0 \\
    \midrule
Total &  &  &  &  &  &  &  &  &  &  &  &  &  &  &  & 6,418.93 & 44.7\% &  & 23.970 & 34.3 \\
    \bottomrule
  \end{tabularx}
  \label{t:convfirstnet-nano-speed}
\end{sidewaystable}

\begin{sidewaystable}[h]
  \scriptsize
  \caption{\textbf{ConvFirstNet-Small} speed projection.}
  \begin{tabularx}{\textwidth}{lrrrrrrrrrr|rrrrrr|rXXX}
    \toprule
    &   &        &    &   &     &    &   &   &     &    & \multicolumn{6}{c|}{Layer OPs [millions]}  &  & & \\
    Block     & N   & s &  H  & W & Ks  & gw & t & C &  R  &  K & Conv & Exp & Prj & Sqz Exc & Depth & Total & \% Peak & Est. \%Peak & Latency [ms] & TFLOP/s \\
    \midrule
Stem & 128 & 2 & 256 & 256 & 3 & 0 & 0 & 3 & 0 & 32 & 28.31 & 0.00 & 0.00 & 0.00 & 1 & 28.31 & 0.0\% & 75.0\% & 0.063 & 57.5 \\
ConvFirst & 128 & 1 & 128 & 128 & 3 & 8 & 3 & 32 & 96 & 32 & 75.50 & 100.66 & 100.66 & 0.00 & 2 & 553.65 & 46.6\% & 0.0\% & 1.983 & 35.7 \\
ConvFirst & 128 & 2 & 128 & 128 & 3 & 8 & 6 & 32 & 192 & 64 & 75.50 & 100.66 & 100.66 & 0.00 & 1 & 276.82 & 0.0\% & 61.0\% & 0.758 & 46.8 \\
ConvFirst & 128 & 1 & 64 & 64 & 3 & 8 & 6 & 64 & 384 & 64 & 37.75 & 201.33 & 201.33 & 0.00 & 4 & 1,761.61 & 76.2\% & 0.0\% & 3.858 & 58.4 \\
ConvFirst & 128 & 2 & 64 & 64 & 3 & 8 & 6 & 64 & 384 & 96 & 37.75 & 100.66 & 75.50 & 0.00 & 1 & 213.91 & 0.0\% & 55.0\% & 0.650 & 42.2 \\
ConvFirst & 128 & 1 & 32 & 32 & 3 & 8 & 6 & 96 & 576 & 96 & 14.16 & 113.25 & 113.25 & 0.00 & 5 & 1,203.24 & 68.7\% & 0.0\% & 2.923 & 52.7 \\
MBConv & 128 & 2 & 32 & 32 & 3 & 8 & 4 & 96 & 384 & 256 & 56.62 & 75.50 & 50.33 & 0.04 & 1 & 182.49 & 0.0\% & 40.8\% & 0.746 & 31.3 \\
MBConv & 128 & 1 & 16 & 16 & 3 & 8 & 4 & 256 & 1024 & 256 & 37.75 & 134.22 & 134.22 & 0.26 & 17 & 5,209.59 & 51.0\% & 0.0\% & 17.047 & 39.1 \\
MBConv & 128 & 2 & 16 & 16 & 3 & 8 & 4 & 256 & 1024 & 256 & 37.75 & 134.22 & 33.55 & 0.26 & 1 & 205.78 & 0.0\% & 41.1\% & 0.835 & 31.5 \\
MBConv & 128 & 1 & 8 & 8 & 3 & 8 & 4 & 256 & 1024 & 256 & 9.44 & 33.55 & 33.55 & 0.26 & 17 & 1,305.74 & 51.4\% & 0.0\% & 4.239 & 39.4 \\
    \midrule
Total  &  &  &  &  &  &  &  &  &  &  &  &  &  &  &  & 10,941.14 & 55.2\% &  & 33.102 & 42.3 \\
    \bottomrule
  \end{tabularx}
  \label{t:convfirstnet-nano-speed}
\end{sidewaystable}

\begin{table}[h]
  \scriptsize
  \caption{\textbf{ConvFirstNet speed projection.} For each of the network sizes, we copy the estimates for the arithmetic utilization (i.e., \% Peak) and performance (TFLOP/s) from the previous tables and use it to estimate the latency of the whole network's latency by dividing MACs by (half) of TFLOP/s. This estimate assumes that the layers left out of the previous tables, namely the layers of the classifier head, have the same average performance as the rest of the network. The head is small compared to the rest of the network, and the bulk of it is a point-wise convolution with a large channel count, which likely could be performed more efficiently than the rest of the network. Therefore, we are likely underestimating network speed slightly by assuming the classifier head runs at the same speed as the rest of the network.}
  \begin{tabularx}{\textwidth}{lrrrXrXXXXXXr}
    \toprule
Model & Pars[M] & MACs[B] & Top1[\%] & Input Size & Batch & Batch MACs[B] & Device TFLOP/s & \% Peak & TFLOP/s & Batch Time[ms] & FPS \\
  \midrule
  ConvFirst-P & 5.91 & 0.86 & 79.858 & 256 & 128 & 110.1 & 76.7 & 47.23\% & 36.2 & 6.078 & 21,060.03 \\
  ConvFirst-N & 10.17 & 1.812 & 81.834 & 256 & 128 & 231.9 & 76.7 & 44.64\% & 34.2 & 13.549 & 9,446.95 \\
  ConvFirst-T & 17.24 & 3.227 & 82.752 & 256 & 128 & 413.1 & 76.7 & 44.69\% & 34.3 & 24.101 & 5,311.04 \\
  ConvFirst-S & 28.55 & 5.493 & 83.312 & 256 & 128 & 703.1 & 76.7 & 55.16\% & 42.3 & 33.238 & 3,851.04 \\
  \bottomrule
  \end{tabularx}
  \label{t:convfirstnet-nano-speed}
\end{table}

\FloatBarrier

\section{Amdahl's Roofline}
\label{a:amdahls-roofline}

How much can we improve the minimum attainable latency of a sequence of parallel kernels by reducing the DRAM transfers of a single kernel?

As in Section~\ref{s:waterline}, ${\mathscr{R}}$ is the peak arithmetic throughput, $\mathscr{B}$ is the peak DRAM memory bandwidth, $n_i$ and $b_i$ are the number of operations and bytes transferred by kernel $i$, and $\frac{n_i}{b_i}$ is its operational intensity.

If kernel $j$ is memory-bound, then $\frac{n_j}{b_j} < \frac{\mathscr{R}}{\mathscr{B}}$, and its
minimum latency is $t_j = \frac{b_j}{\mathscr{B}}$.
By reducing the bytes transferred to $\hat{b}_j = \frac{n_j}{\mathscr{R} / \mathscr{B}}$ or less, the kernel becomes compute bound with new minimum latency $\hat{t}_j = \frac{n_j}{\mathscr{R}}$.
Therefore, the maximum improvement equals
\begin{equation}
\begin{split}
  \frac{T}{\hat{T}} &= \frac{T}{T - t_j + \hat{t}_j} \\
  &= \frac{1}
          {1 - \frac{t_j}
                              {T}
                       + \frac{\hat{t}_j}
                              {T}
          } \\
  &= \frac{1}{1 - f_B + f_R}
  \end{split}
\label{e:amdahl-roofline}
\end{equation}
where
\begin{equation*}
  f_B = \frac{b_j}{\mathscr{B} T}\text{,} \quad \quad f_R = \frac{n_j}{\mathscr{R} T}
\end{equation*}
are the minimum latency of the memory bound and compute bound versions of kernel $j$, measured as a fraction of the total minimum attainable latency before optimization,
\begin{equation*}
  T = \sum_i{t_i}~.
\end{equation*}
Equation~\eqref{e:amdahl-roofline} can be understood as a variation of Amdahl's law~\cite{amdahl1967validity} applied to roofline analysis~\cite{williams2009roofline}.

If kernel $j$ is already compute bound, then further reducing its DRAM transfers has no effect on minimum latency. Yet it does increase the mediant operational intensity (Equation~\ref{e:mediant-opint}). This is the reason why the roofline performance model is not correct for a sequence of kernels.
\end{document}